# REACT-LLM: A Benchmark for Evaluating LLM Integration with Causal Features in Clinical Prognostic Tasks

Linna Wang[1*], Zhixuan You[1*], Qihui Zhang[2*], Jiunan Wen[1], Ji Shi[1], Yimin Chen[3], Yusen Wang[1], Fanqi Ding[1], Ziliang Feng[1], Li Lu[1†]

[1]Sichuan University
[2] Peking University
[3] The Second Affiliated Hospital of Kunming Medical University

lenawang@stu.scu.edu.cn, youzhixuan@stu.scu.edu.cn, zqhui_scu@foxmail.com, jiunanwen@stu.scu.edu.cn, 2024141520215@stu.scu.edu.cn, 20241105@kmmu.edu.cn, 2024141520213@stu.scu.edu.cn, dingfanqi@stu.scu.edu.cn, fengziliang@scu.edu.cn, luli@scu.edu.cn

## Abstract

Large Language Models (LLMs) and causal learning each hold strong potential for clinical decision making (CDM). However, their synergy remains poorly understood, largely due to the lack of systematic benchmarks evaluating their integration in clinical risk prediction. In real-world healthcare, identifying features with causal influence on outcomes is crucial for actionable and trustworthy predictions. While recent work highlights LLMs' emerging causal reasoning abilities, there lacks comprehensive benchmarks to assess their causal learning and performance informed by causal features in clinical risk prediction. To address this, we introduce REACT-LLM, a benchmark designed to evaluate whether combining LLMs with causal features can enhance clinical prognostic performance and potentially outperform traditional machine learning (ML) methods. Unlike existing LLM-clinical benchmarks that often focus on a limited set of outcomes, REACT-LLM evaluates 7 clinical outcomes across 2 real-world datasets, comparing 15 prominent LLMs, 6 traditional ML models, and 3 causal discovery (CD) algorithms. Our findings indicate that while LLMs perform reasonably in clinical prognostics, they have not yet outperformed traditional ML models. Integrating causal features derived from CD algorithms into LLMs offers limited performance gains, primarily due to the strict assumptions of many CD methods, which are often violated in complex clinical data. While the direct integration yields limited improvement, our benchmark reveals a more promising synergy: LLMs serve effectively as knowledge-rich collaborators for identifying and optimizing causal features. Additionally, in-context learning improves LLM predictions when prompts are tailored to the task and model. Different LLMs show varying sensitivity to structured data encoding formats, for example, open-source models perform better with JSON, while smaller models benefit from narrative serialization. These findings highlight the need to match prompts and data formats to model architecture and pretraining.

**Code** —
https://github.com/LinnaWang-Lena/REACT_LLM

**Extended version** — https://arxiv.org/abs/2511.07127

---

[*]These authors contributed equally.
[†]Corresponding author.

## Introduction

In clinical environments such as the Intensive Care Unit (ICU), timely and accurate risk assessment is critical for enabling early interventions and improving patient outcomes (Alizadehsani et al. 2021; Placido et al. 2023; Fihn et al. 2024; Yeh et al. 2024). The widespread adoption of Electronic Health Records (EHRs) has provided access to rich, structured clinical data, opening new opportunities for data-driven decision support (Nguyen et al. 2021; Khalifa, Albadawy, and Iqbal 2024; Wang et al. 2025). One key application is prognostic modeling (Van Smeden et al. 2021), which estimates the risk of patients developing specific conditions over time. In this context, machine learning (ML) models outperform traditional statistical methods in terms of scalability and predictive performance. However, ML models often depend on high-dimensional features and remain vulnerable to input variability and spurious correlations (Rajpurkar et al. 2022; Wang et al. 2025). This has motivated growing interest in applying causal discovery (CD) to clinical downstream tasks, as advances in CD methods (Zhou and Chen 2022; Lagemann et al. 2023; Zanga, Ozkirimli, and Stella 2022; Feuerriegel et al. 2024) enable the extraction of meaningful causal relationships from observational data.

Recently, Large Language Models (LLMs), known for their impressive capabilities across a wide range of natural language processing tasks (Van Veen et al. 2024; Van Sonsbeek et al. 2023), have been increasingly applied to clinical applications (Thirunavukarasu et al. 2023; Ferdush, Begum, and Hossain 2024; Singhal et al. 2025; Van Veen et al. 2024; Liu et al. 2023; Sandmann et al. 2024; Kafkas et al. 2025; Kang et al. 2025). Leveraging strategies such as fine-tuning (Ben Shoham and Rappoport 2024; Wang et al. 2024a), retrieval-augmented generation (RAG) (Bedi, Thukral, and Dhiman 2025), and task-specific prompting (Zheng et al. 2025), LLMs have shown promise in clinical risk prediction. For example, (Ben Shoham and Rappoport 2024) fine-tuned a pre-trained LLM to predict diagnoses and hospital readmissions, achieving state-of-the-art performance.

Although LLMs and causal learning each show strong potential for CDM, their potential synergy remains largely unexplored. This is primarily due to the lack of a systematic benchmark for evaluating their integrated application in clinical risk prediction tasks. Moreover, evidence on LLM performance in this domain is not uniformly positive, with some studies concluding that LLMs are not yet prepared for autonomous CDM (Hager et al. 2024; Brown et al. 2025). A recent benchmark, ClinicalBench (Chen et al. 2024), systematically evaluated general-purpose and medical-specific LLMs against traditional ML models on 3 prognostic tasks. The study revealed that despite variations in model scale and the use of different prompting or fine-tuning strategies, LLMs consistently underperformed ML models. Therefore, the predictive capabilities of LLMs in clinical risk prediction require further investigation. A comprehensive benchmark that integrates LLMs, causal learning, and clinical prediction is urgently needed. Motivated by this, this study aims to rigorously evaluate whether incorporating causal knowledge can improve the performance of LLMs on critical clinical risk prediction tasks. Unlike prior work limited to narrow clinical endpoints, we extend the evaluation to a broader set of prognostic outcomes to address the central question: *Can the integration of LLMs with causal features enhance performance in clinical prognostic tasks and potentially outperform traditional ML models?*

**REACT-LLM Design.** To answer this question, we present REACT-LLM (Risk Evaluation and Causal features Test with LLMs), a novel benchmark for assessing how effectively LLMs can serve as inference and error-correction experts that augment CD methods in uncovering causal features of clinical risk outcomes (Figure 1). Using structured data from 2 real-world datasets, MIMIC-III(Johnson et al. 2016) and MIMIC-IV(Johnson et al. 2023), we investigate 7 representative prognostic tasks: (1) In-hospital mortality (InHospDeath), (2) 30-day hospital readmission (Readmit30), (3) Multiple ICU stays during a single hospitalization (MultiICU), (4) Sepsis during ICU stay (SepsisICU), (5) Acute kidney injury during ICU stay (AKIICU), (6) Prolonged hospital stay (LOS), and (7) Early ICU admission (EarlyICU). Here are the main tasks:

▷ Baseline evaluation: Benchmark all MLs/LLMs on 7 outcomes across 2 datasets using complete feature sets.

▷ Prompt engineering evaluation: Beyond direct prompting, 4 representative prompt engineering strategies (Chain-of-Thought (CoT), Self-Reflection (SR), Role-Playing (RP), and In-Context Learning (ICL)) are employed to assess LLMs performance across 7 tasks.

▷ Input format sensitivity evaluation: Assess LLMs performance across 5 formats for structured patient data: Row-Column Format (RCF), JSON, LaTeX, Template-Based Natural Language (TBNL) and Narrative Serialization (NS).

▷ Causal feature evaluation: Use 3 representative CD methods to identify features with direct or indirect causal relationships to each outcome. Evaluate LLMs using only these causal features as input.

▷ LLM-assisted causal feature editing evaluation: Prompt LLMs to optimize the causal feature sets derived from 3 CD methods, and separately, to generate causal feature sets for each outcome relying solely on their internal knowledge.

Overall, the contributions can be summarized as follows:

- ⊞ Benchmark: Evaluate **3** CD algorithms, **6** ML models, and **15** LLMs (spanning diverse model sizes and architectures) for predicting **7** clinical outcomes.
- 🗄 Two Datasets: We curate the MIMIC-III and MIMIC-IV datasets, encompassing 6 categories of clinical information and 7 prognostic outcome labels, providing a comprehensive foundation for evaluating LLM-based medical risk prediction models.
- 💡 Actionable Insights & Nuanced Findings: (1) LLMs have yet to outperform traditional ML models in clinical outcome prediction. (2) Engineering prompts offer slight gains under imbalanced EHR conditions. (3) LLMs refine CD-derived features effectively, enabling a human–AI synergy for causal feature engineering. (4) Different LLMs exhibit varying sensitivity to structured EHR encoding formats. RCF benefits proprietary large models, JSON consistently enhances performance in open-source models, while NS proves effective for smaller models. LLMs can infer causal relations even among variables with ambiguous or coded names, such as ICD codes.

## REACT-LLM Construction

### Goals

- **Causal Feature Identification.** Recover the causal structure from observational data using 3 CD methods by assigning each outcome to its direct and indirect causes.
- **Clinical Binary Outcome Prediction.** Estimate the probability of binary outcomes.
- **LLM Evaluation Tasks.** Evaluate LLMs across 5 groups of experiments to address the following questions:

  ❏ Q1: Can LLM outperform traditional ML in clinical prognosis? → *Based on baseline evaluation.*

  ❏ Q2: Can prompt engineering boost LLM performance in clinical risk prediction? → *Based on prompt engineering evaluation.*

  ❏ Q3: How does the encoding format of structured EHR data affect LLM performance? → *Based on input format sensitivity evaluation.*

  ❏ Q4: Can CD improve LLM predictions by identifying causal features? → *Based on causal feature evaluation.*

  ❏ Q5: Can LLM validate or identify causal features in clinical contexts? → *Based on LLM-assisted causal feature editing evaluation.*

### Datasets Preprocessing

**Study Population.** This study uses 2 public datasets[1]: MIMIC-III (v1.4) (Johnson et al. 2016) and MIMIC-IV (v3.1) (Johnson et al. 2023), which contain de-identified EHRs from ICU and emergency department admissions at Beth Israel Deaconess Medical Center in Boston. MIMIC-III contains ICU records from 2001 to 2012 and is widely used in critical care research, while MIMIC-IV extends coverage through 2022 with an updated schema and improved data quality. We include only adult patients (age $\geq 18$), retained the first ICU stay per admission, and excluded ICU stays shorter than one day to ensure clinical relevance.

---
[1]https://physionet.org

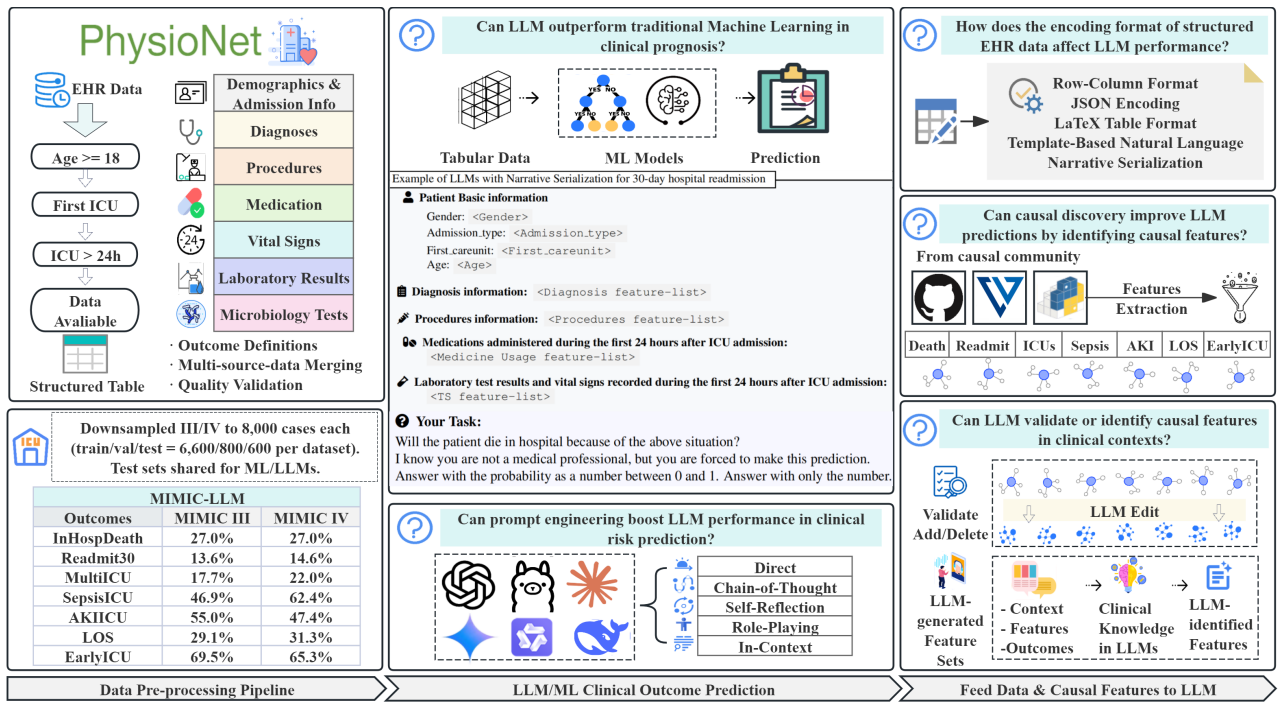

Figure 1: REACT-LLM framework diagram. The *Data Preprocessing Pipeline* outlines data processing steps, 7 outcome labels, and dataset splits. The MIMIC-LLM table presents our curated data, showing the proportion of positive samples for each outcome. *LLM/ML Clinical Outcome Prediction* presents 2 prediction tasks and a prompt sample. *Feed Data & Causal Features to LLM* illustrates 3 evaluation tasks.

**Data collection.** We extract 5 categories of features to characterize each ICU admission: (1) demographics and admission details (age, gender, admission type, initial ICU unit); (2) 65 high-frequency admission diagnoses from MIMIC-III/IV, encoded as binary indicators (present/absent); (3) 27 high-frequency clinical procedures (binary encoded); (4) 55 medications administered within the first 24 hours (total dosage per drug); and (5) 115 vital signs and laboratory results from the first 24 hours (median summarized). The 24-hour window supports early prediction while capturing key clinical information from the initial phase of ICU care.

## Experiment Setup

**Prediction Tasks.** We define 7 outcome labels (not mutually exclusive): (1) InHospDeath: in-hospital mortality; (2) Readmit30: a binary indicator set to 1 if a subsequent hospital admission occurs within 30 days of discharge; (3) MultiICU: more than one ICU stay associated with the same hospital admission; (4) SepsisICU: identified using the database-provided label based on the Sepsis-3 definition (Singer et al. 2016), which requires a SOFA score $\geq 2$ and clinical suspicion of infection; (5) AKIICU: defined according to KDIGO criteria (Khwaja 2012), using creatinine or urine output when available, with baseline creatinine taken as the lowest value within the prior 7 days; (6) LOS: length of hospital stay exceeding 14 days; and (7) EarlyICU: ICU admission occurring within 12 hours of hospital admission.

**Dataset Division.** To manage the computational cost of querying LLMs, we apply stratified random sampling to ensure proportional representation of positive cases for each label, selecting 8,000 samples from each datasets, resulting in a combined cohort of 16,000 patients. For traditional ML models, the combined dataset is split into training (6,600), validation (800), and test (600) sets. The test sets from both datasets are shared with LLM evaluations to ensure fair and consistent comparisons.

**Metrics.** This paper uses AUROC, AUPRC and F1 score. We report 95% confidence intervals based on 1,000 bootstrap samples for ML models and 5-run results for LLMs.

## Method

**Causal Discovery Models:** We evaluate 3 representative CD approaches: (1) functional-based (DirectLiNGAM (Shimizu et al. 2011)), (2) score-based (GES (Chickering 2002)), and (3) gradient-based (CORL (Wang et al. 2021)). All methods were implemented using the `gCastle` toolkit[2], an open-source causal discovery library developed by Noah's Ark Lab (Zhang et al. 2021). We apply each CD method 5 times on the full dataset to identify direct and indirect causes for each outcome. Features appearing in more than 2 runs are retained to form the causal feature set. The parameter Settings are shown in the Appendix A.2.

---
[2]https://github.com/huawei-noah/trustworthyAI/tree/master/gcastle

**Benchmarked ML Models:** We include 6 common used baseline models: AdaBoost, Decision Tree (DT), Logistic Regression (LR), Random Forest (RF), Support Vector Machine (SVM) and XGBoost. All models are run with default hyperparameters to ensure fairness and reproducibility, allowing us to examine whether the LLM can outperform unoptimized traditional ML models.

**Prompt Protocols.** Following predefined clinical categories (patient demographics, diagnoses, procedures, medications, and laboratory results including vital signs), we evaluate 5 prevalent encoding strategies for transforming structured EHR data into formats suitable for LLM input: (1) RCF: Encodes records in a flat, CSV-like format using comma-separated values. (2) JSON Encoding: Represents data as hierarchical key-value pairs organized by clinical categories. (3)LaTeX Table Format: Structures EHR data into formatted LaTeX-style tables by category. (4) TBNL: Converts each clinical category into sentence-level descriptions using predefined templates to enhance textual fluency. (5) NS: Converts tabular EHR data into coherent, human-readable narratives, structured by the predefined clinical categories. We proceed as follows:

- Baseline Evaluation: Applies the Direct Prompting approach using LLMs on input formatted with NS.
- Prompt Engineering Evaluation: Extends beyond Direct Prompting by incorporating strategies CoT, SR, RP and ICL, all under the NS format.
- Input Format Sensitivity Evaluation: Tests the Direct Prompting approach across all 5 input formats (RCF, JSON, LaTeX, TBNL, NS) to assess the impact of encoding on performance.
- Causal Feature Evaluation: Uses only the feature subsets identified by CD methods, test under the Direct Prompting setting with NS input.

The prompt protocols for these 4 groups follow a consistent design and are detailed in the Appendix C and E.

For the LLM-assisted causal feature editing evaluation, we design 2 protocols (details are in the Appendix D):

- LLM edited CD feature sets: For each clinical outcome, feature sets generated by a CD algorithm are provided to the LLM, which was prompted to act as an expert in intensive care medicine and refine the list. The LLM's output was used as the optimized feature set.
- LLM-Generated causal feature sets: The LLM independently generates a causal feature set based solely on its internalized clinical knowledge.

**LLM Implement details:** Experiments are conducted on 3 NVIDIA RTX A800 GPUs. We benchmark a diverse set of LLMs spanning different architectures and scales. To ensure reproducibility, all inferences use greedy decoding (temperature = 0, do_sample = False). *Proprietary models* included advanced reasoning models (GPT-o1, GPT-o3 mini, Claude-3.7-Sonnet, Gemini-2.5-Pro, Gemini-2.5-Flash), large models (GPT-4o, Claude-4, Claude-3.5-Haiku), and small models (GPT-4o-mini). *Open-source models* included a top reasoning model (DeepSeek-R1), large models (Llama-3.1-405b, DeepSeek-V3, Qwen3-235b), and small models (Qwen3-8b, Qwen3-14b). Details are in Appendix A.1.

## Empirical Results and Analysis

### Baseline evaluation

We evaluate 15 LLMs with direct prompting across 7 clinical prediction tasks on 2 datasets, using 3 metrics (complete results in the Appendix B.1 and B.2). As shown in Table 1, among traditional ML models, XGBoost, RF, and LR consistently achieve top performance across clinical outcomes. In the open-source LLM category, larger models (e.g., DeepSeek-V3, Llama-3.1-405b) outperform smaller and thinking models more frequently. While thinking models rarely lead on individual tasks, they exhibit more stable performance than smaller models overall. A similar trend holds for proprietary LLMs: larger models generally outperform other ones. Notably, Gemini-2-Flash achieves the highest AUPRC on the EarlyICU task (0.8540), exceeding the best ML model (RF, 0.8473). Thinking models in this group also show more consistent performance than smaller variants. Overall, proprietary LLMs outperform open-source ones more often across tasks.

However, with the exception of Gemini-2-Flash on the EarlyICU AUPRC metric, none of the LLMs surpass traditional ML models on any clinical prediction task. In most cases, LLM performance lags behind ML baselines by a substantial margin, typically around 10–20%.

Based on this, we select **6** representative small, large and thinking models from both open-source and proprietary LLMs for subsequent experiments.

✔ Finding for Q1: Current LLMs remain immature and unreliable for clinical prognostic decision support. Among LLMs, proprietary and large-scale models tend to offer relatively better performance.

### Prompt engineering evaluation

We evaluate 6 LLMs with 5 prompting strategies across 7 clinical prediction tasks(complete results in Appendix B.3). Across all tasks and metrics, ICL surpasses the baseline 41 times and ranks best in 22. SR also performs well, exceeding the baseline in 38 cases and leading in 19. Role-Playing and CoT follow. As shown in Figure 2, ICL outperforms direct prompting in 18 cases and achieves the highest AUROC in 13. Prompting strategies yield clear improvements on the Readmit30 task. ICL enhances performance: Gemini-2-Pro's AUROC increases by 0.077, GPT-o3-mini's AUROC and F1 improve by 0.072 and 0.084, respectively, and Qwen3-235B gains 0.059 in AUROC. DeepSeek-R1 benefits more from CoT prompting, with a 0.0271 increase in AUPRC and a 0.0912 gain in AUROC. However, despite these gains, none of the prompting strategies enable LLMs to surpass traditional ML models across any outcome.

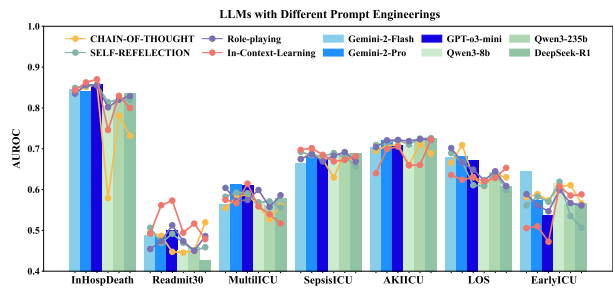

Figure 2: AUROC of LLMs with Different Engineering Strategies on MIMIC-III across 7 Labels. Bars indicate baseline scores obtained via direct prompting.

| Outcome | InHospDeath | | Readmit30 | | MultiICU | | SepsisICU | | AKIICU | | LOS | | EarlyICU | |
|---|---|---|---|---|---|---|---|---|---|---|---|---|---|---|
| Positive Ratio | 27.0% | | 13.6% | | 17.7% | | 46.9% | | 55.0% | | 29.1% | | 69.5% | |
| Metric | ROC | PRC | ROC | PRC | ROC | PRC | ROC | PRC | ROC | PRC | ROC | PRC | ROC | PRC |
| **Machine Learning Models** | | | | | | | | | | | | | | |
| SVM | 0.8949 | 0.7707 | 0.6668 | 0.2393 | 0.7467 | 0.4141 | 0.8059 | 0.7956 | 0.8221 | 0.8558 | 0.8499 | 0.6949 | **0.7336** | 0.8304 |
| LR | 0.9079 | 0.8047 | **0.7326** | **0.2912** | **0.7555** | **0.4419** | 0.8059 | 0.7856 | 0.8285 | 0.865 | **0.8552** | 0.7022 | 0.6884 | 0.8406 |
| DT | 0.7022 | 0.6354 | 0.5323 | 0.2535 | 0.5617 | 0.3227 | 0.659 | 0.7265 | 0.6499 | 0.7815 | 0.6482 | 0.5699 | 0.5885 | 0.8347 |
| RF | 0.8936 | 0.7806 | 0.6222 | 0.2122 | 0.7109 | 0.3507 | **0.8229** | **0.8059** | 0.8323 | 0.8576 | 0.8521 | 0.6917 | 0.7263 | **0.8473** |
| Adaboost | 0.8694 | 0.7231 | 0.6837 | 0.253 | 0.7418 | 0.3532 | 0.8111 | 0.795 | 0.8318 | **0.8655** | 0.8404 | 0.6736 | 0.7180 | 0.8443 |
| XGboost | **0.9088** | **0.8148** | 0.6863 | 0.2469 | 0.7199 | 0.3737 | 0.8183 | 0.7964 | **0.8332** | 0.8606 | 0.8516 | **0.7099** | 0.7312 | 0.8319 |
| **Open-source Large Language Models** | | | | | | | | | | | | | | |
| Qwen3-14b | 0.8075 | 0.6544 | 0.4502 | 0.1002 | 0.5719 | 0.1472 | 0.6848 | 0.6528 | 0.7092 | 0.7679 | 0.6219 | 0.3143 | 0.6179 | 0.739 |
| Qwen3-8b | 0.8076 | 0.6537 | 0.4511 | 0.0999 | 0.5711 | 0.1476 | 0.6866 | 0.6546 | 0.7089 | 0.7676 | 0.6172 | 0.3129 | 0.615 | 0.7377 |
| Llama-3.1-405b | 0.7012 | 0.4658 | 0.4929 | 0.1304 | 0.5545 | 0.1681 | 0.5687 | **0.7139** | 0.6988 | **0.7969** | 0.5355 | **0.6419** | 0.5389 | 0.7477 |
| Qwen3-235b | 0.8194 | 0.6878 | 0.4580 | 0.0964 | 0.5615 | 0.1695 | **0.6894** | 0.6604 | 0.7221 | 0.7409 | 0.6425 | 0.5511 | 0.5641 | 0.7197 |
| DeepSeek-V3 | 0.8370 | 0.6873 | 0.4478 | 0.0796 | 0.5845 | 0.2126 | 0.6756 | 0.6384 | **0.7413** | 0.7862 | 0.6531 | 0.4192 | 0.5567 | 0.7796 |
| DeepSeek-R1 | 0.8363 | 0.6649 | 0.4285 | 0.1030 | 0.5793 | 0.2104 | 0.6883 | 0.6487 | 0.7270 | 0.7667 | 0.6096 | 0.3732 | 0.5663 | 0.7660 |
| **Proprietary Large Language Models** | | | | | | | | | | | | | | |
| GPT-4o-mini | 0.8080 | 0.7036 | 0.4712 | 0.091 | 0.5538 | 0.1748 | 0.6673 | 0.6479 | 0.7104 | 0.7487 | 0.6158 | 0.4195 | 0.5450 | 0.7855 |
| Gemini-2-Flash | 0.8475 | 0.6817 | 0.4881 | 0.1209 | 0.5634 | 0.1759 | 0.6653 | 0.6703 | 0.7045 | 0.7507 | 0.6799 | 0.4592 | **0.6455** | **0.8540** |
| GPT-4o | 0.8223 | 0.6961 | 0.4780 | **0.2133** | 0.5790 | 0.1878 | 0.6751 | 0.6298 | 0.7069 | 0.7687 | 0.6618 | 0.4375 | 0.5513 | 0.7676 |
| Claude-3.5-Haiku | 0.8149 | 0.7020 | **0.5227** | 0.0992 | 0.5517 | 0.1616 | 0.6586 | 0.5482 | 0.7085 | 0.7501 | 0.6172 | 0.4412 | 0.5970 | 0.7640 |
| Claude-4 | 0.8252 | 0.7039 | 0.4500 | 0.0860 | 0.5689 | 0.2167 | 0.6707 | 0.6765 | 0.6846 | 0.7449 | 0.6354 | 0.5465 | 0.5128 | 0.7559 |
| Gemini-2-Pro | 0.8424 | 0.6904 | 0.4848 | 0.1177 | 0.6135 | **0.2300** | 0.6806 | 0.6373 | 0.7214 | 0.7625 | **0.7122** | 0.5089 | 0.5748 | 0.7722 |
| GPT-o1 | 0.8396 | 0.6695 | 0.4742 | 0.1163 | **0.6194** | 0.2269 | 0.6605 | 0.5984 | 0.7127 | 0.7558 | 0.6331 | 0.3691 | 0.5931 | 0.7675 |
| GPT-o3-mini | 0.8585 | 0.6831 | 0.5015 | 0.1224 | 0.6109 | 0.1955 | 0.6758 | 0.6585 | 0.7095 | 0.7433 | 0.6712 | 0.4735 | 0.5408 | 0.8013 |
| Claude-3.7-Sonnet | **0.8699** | **0.7365** | 0.4665 | 0.1146 | 0.5815 | 0.1999 | 0.6530 | 0.6101 | 0.7139 | 0.7588 | 0.6891 | 0.4883 | 0.5867 | 0.7544 |

Table 1: Baseline performance of LLMs and traditional ML models across 7 clinical prediction tasks on MIMIC-III. Results on MIMIC-IV are provided in Appendix B.1 and B.2. 'Positive Ratio' refers to the proportion of samples labeled as 1. 'ROC' denotes AUROC, and 'PRC' denotes AUPRC. Bold scores indicate the best performance within ML models and LLMs category.

✔ Finding for Q2: ICL can enhance LLM performance in clinical prediction tasks. In highly imbalanced settings, prompting strategies show some benefits. However, no single strategy consistently improves results across all tasks and models, suggesting their effectiveness depends on the clinical context and model capacity.

### Input format sensitivity evaluation

We analyze the performance of 4 LLMs across 3 metrics using 5 input format strategies (complete results in the Appendix B.4). As shown in Figure 3, the proprietary large model GPT-o3-mini responds best to RCF, with weaker performance on NS and TBNL. The open-source large model Qwen3-235b performs best with JSON, leading across all metrics and outperforming RCF by up to 0.064, but responds least effectively to RCF. Among smaller models, the proprietary Gemini-2-Flash performs best with NS, achieving its highest AUROC and AUPRC, surpassing RCF by 0.051 and 0.026, respectively. The open-source Qwen3-8b shows strong results with both JSON and NS.

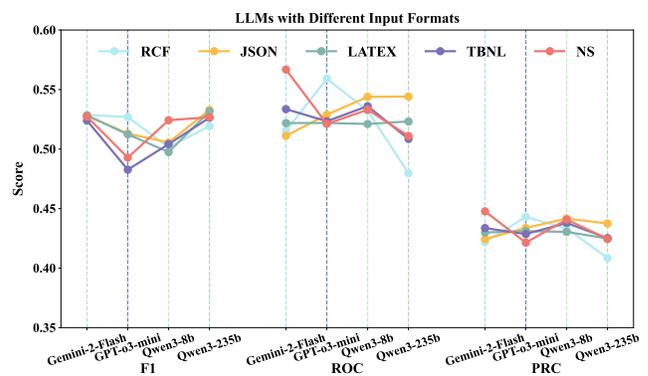

Figure 3: Average performance of LLMs with 5 Input Formats on MIMIC-III. All 3 scores represent the mean values across 7 outcomes.

✔ Finding for Q3: Different LLMs show varying sensitivity to structured EHR encoding formats. RCF works well for proprietary large models. JSON consistently improves performance for open-source models. LaTeX tables offer limited benefit across models. TBNL produces mixed results, indicating rigid templates require task-specific tuning. NS is particularly effective for smaller models.

Why do different types of LLMs prefer different input formats? This preference likely stems from the interplay between pretraining data, model architecture, and capacity. Proprietary large models tend to favor RCF, possibly due to exposure to structured documents during pretraining. This aligns with previous study (Li et al. 2024), which suggest LLMs exhibit a bias toward formal data formats. In contrast, open-source large models prefer JSON, as their training data often include public datasets rich in JSON-formatted templates, API references, and structured records, common on platforms like Hugging Face. JSON's key-value structure closely matches the data distribution these models encounter, and a study (Zhu et al. 2024) shows that reinforcing JSON format improves hierarchical parsing. Smaller models perform better with NS, which converts inputs into fluent narrative text. This format reduces reliance on structure parsing and better matches the natural language data these models are trained on, aligning with a previous study showing that smaller models benefit from training-aligned data distributions (Yam and Paek 2024).

### Causal feature evaluation

In this section, we investigate our core hypothesis (Q4): can causally selected features improve LLM prediction? We evaluate the performance of 6 LLMs using 3 CD feature sets (complete results in the Appendix B.5). As shown in Figure 4, across all LLMs, CD-derived features often lead to performance degradation. Compared to the baseline, DirectLiNGAM leads to the most significant drop. However, CORL yields slight improvements on Gemini-2-Pro, DeepSeek-R1 and Qwen3-235b. GES shows consistent gains on open-source models including DeepSeek-R1, Qwen3-8b and Qwen3-235b.

Contrary to expectations, although some improvements are observed, LLMs leveraging CD features do not consistently outperform the baseline. However, this does not reflect the quality of the CD methods themselves, as ground-truth causal graphs are rarely available in clinical data for validation. In clinical risk prediction, strict assumptions in CD algorithms may result in a limited set of outcome-related causal features, while non-causal but highly correlated features still provide strong predictive signals. Additionally, most CD methods lack prior knowledge and domain-specific constraints, making it difficult to recover complex causal structures from high-dimensional clinical data.

✔ Finding for Q4: Causal feature subsets derived from commonly used CD algorithms did not consistently improve LLMs prediction. This may be due to strict assumptions inherent in many CD methods (e.g., the Causal Faithfulness assumption or the absence of hidden confounders), which are often violated in complex, high-dimensional clinical data. However, this highlights the potential of further exploring whether integrating LLM-derived prior knowledge into CD outcomes can enhance performance (Zhou et al. 2024; Ban et al. 2025).

### LLM-assisted causal feature editing evaluation

Building on the findings from Q4, we further explore whether LLM-assisted causal features could enhance LLM performance. We evaluate 2 strategies: (1) optimizing causal feature sets derived from CD methods using different LLMs, and (2) allowing LLMs to directly generate causal feature sets based on clinical knowledge. In both cases, the corresponding LLM is used to predict the outcome using its respective feature set (complete results in the Appendix B.5).

Figure 4 shows the average AUROC performance of each model on both MIMIC-III and MIMIC-IV datasets. LLM optimization is key for DirectLiNGAM feature sets. Except for DeepSeek-R1, the optimized DirectLiNGAM features consistently outperform their unoptimized counterparts, though still fall short of the baseline. For CORL and GES, optimization leads to mixed results, with both improvements and declines. Notably, optimized GES achieves further gains on Qwen3-8b and Qwen3-235b, surpassing the baseline. While LLM-optimized CD features do not always exceed baseline performance, they generally improve upon raw CD outputs. LLM-generated causal features often outperform those from CORL and DirectLiNGAM, and show competitive performance with GES. To further analyze this performance, Table 2 provides a breakdown by dataset for representative proprietary and open-source LLMs, showing that LLM-generated causal feature sets and LLM-optimized GES feature sets yield comparable results.

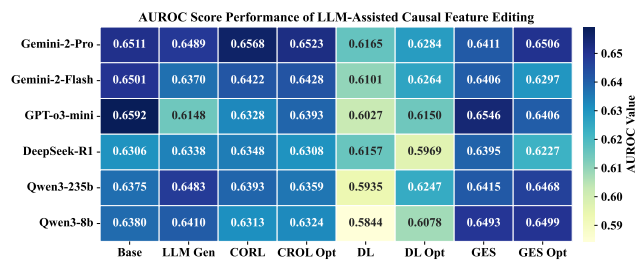

Figure 4: Average AUROC of LLMs across 7 outcomes over MIMIC-III and MIMIC-IV. 'Base' denotes performance using all features. 'LLM Gen' refers to the causal feature set generated by the LLM itself. 'DL' indicates DirectLiNGAM. 'Opt' indicates the optimized feature set refined from the CD outputs by the corresponding LLM.

|  | MIMIC III | | MIMIC IV | |
|---|---|---|---|---|
| Model | AUROC | F1 | AUROC | F1 |
| Gemini-2-Pro | | | | |
| Base | 0.6571 | 0.5471 | 0.6451 | 0.5426 |
| GES Opt | 0.6403 | 0.5142 | 0.6420 | 0.5479 |
| LLM Gen | 0.6559 | 0.5474 | 0.6453 | 0.5673 |
| Qwen3-8b | | | | |
| Base | 0.6368 | 0.5214 | 0.6392 | 0.5456 |
| GES Opt | 0.6449 | 0.4732 | 0.6537 | 0.5240 |
| LLM Gen | 0.6518 | 0.4696 | 0.6481 | 0.5288 |

Table 2: Average performance of LLMs across 7 outcomes on MIMIC-III and MIMIC-IV, respectively. 'Base' denotes performance using all features. 'LLM Gen' denotes the causal feature set generated by the LLM itself, while 'Opt' refers to the GES causal features optimized by the LLM.

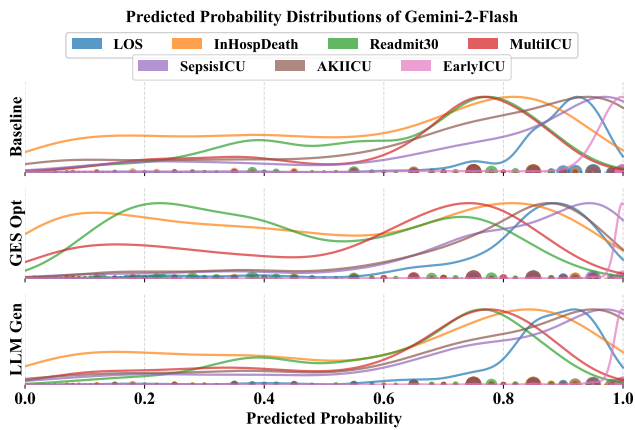

Figure 5: Predicted probability distributions of Gemini-2-Flash across 7 outcomes using 3 different feature sets on MIMIC III: Baseline (all features), Optimized GES (causal features optimized by Gemini-2-Flash), and LLM-Generated (features generated by Gemini-2-Flash).

Figure 5 shows the predicted probability distributions of Gemini-2-Flash using the baseline features, optimized GES features and LLM-generated causal features. For outcomes like Readmit30, MultiICU, and InHospDeath, the LLM-generated features exhibit prediction preferences similar to the baseline, tending toward more severe outcomes. Across all 3 feature sets, the prediction distributions for probabilities above 0.5 are largely consistent.

✔ Finding for Q5: LLMs show potential in optimizing and identifying causal features in clinical settings. This highlights a valuable synergy: CD algorithms provide a structured foundation, while LLMs contribute prior knowledge to refine and enhance feature selection (Darvariu, Hailes, and Musolesi 2024; Zhou et al. 2024). Our results further demonstrate that LLMs can serve as a complementary tool for causal feature selection in clinical prediction tasks, aligning with recent studies showing that LLMs are capable of inferring causal relationships from clinical data (Naik et al. 2024; Kiciman et al. 2023).

## Related works

Clinical decision support is rapidly advancing, driven by the growth of medical data and the shift toward precision medicine. In ICUs, robust and interpretable ML models like LR, XGBoost, and RF are widely used for tasks such as predicting in-hospital mortality (Wang et al. 2025), readmission (Fathy, Emeriaud, and Cheriet 2025), AKI (Lin, Shi, and Kong 2025), and sepsis (Gao et al. 2024), enhancing diagnostic efficiency and enabling early intervention for high-risk patients (Hyland et al. 2020).

Recently, LLMs have shown promise by directly processing unstructured text, preserving critical clinical details often lost in manual feature extraction (Xu et al. 2024; Hager et al. 2024; Wang et al. 2024b; Ahmed et al. 2025). Through fine-tuning and RAG, LLMs integrate hospital-specific data with general medical knowledge, improving decision support (Ansari et al. 2025; Jin et al. 2024). Current benchmarks focus on clinical question answering, multimodal integration, and real-world evaluation frameworks (Liu et al. 2024; Jin et al. 2021; Bae et al. 2023; Budler et al. 2025; Esteitieh, Mandal, and Laliotis 2025).

Meanwhile, causal learning enhances clinical prediction interpretability at both model and data levels. ML may overfit spurious correlations, resulting in unreliable predictions (Huang et al. 2025). In contrast, causal learning aims to uncover the true causal structures, offering more robust and interpretable insights (Richens, Lee, and Johri 2020; Feuerriegel et al. 2024; Zhou and Chen 2022; Zanga, Ozkirimli, and Stella 2022). Major causal discovery methods include constraint-based (e.g., PC (Spirtes, Glymour, and Scheines 2000), score-based (e.g., GES (Chickering 2002)), functional causal models (e.g., DirectLiNGAM (Shimizu et al. 2011)), and optimization-based approaches (e.g., CORL (Wang et al. 2021)). Among them, constraint-based methods are limited by reliance on the faithfulness assumption and inability to determine causal directions within Markov equivalence classes(Zhou and Chen 2022).

Integrating LLMs with causality advances clinical prediction by producing reliable causal chains, enhancing counterfactual reasoning, and translating complex causal relationships into clinically meaningful insights (Zeng et al. 2025; Kiciman et al. 2023; Kweon et al. 2024). However, large-scale evaluations of LLM and causal methods specifically for clinical risk prediction remain scarce. Detailed related work can be found in Appendix F.

## Conclusion

This study evaluates LLMs with causal features in clinical prognostic tasks. While integrating CD features into LLMs did not yield notable gains in clinical risk prediction, this does not diminish the inherent value of CD. The modest performance is largely due to limitations of current CD algorithms in clinical settings: the strict assumptions that often result in sparse or incomplete feature sets. These challenges highlight the need for caution when applying CD outputs directly to LLM-based tasks without further refinement. Nonetheless, our findings suggest a promising synergy: CD provides a data-driven foundation for causal feature identification, while LLMs contribute rich domain knowledge to enhance feature selection. This aligns with growing evidence that LLMs can assist in uncovering causal relationships in complex biomedical contexts. Moving forward, incorporating LLM-guided priors into the CD process, or using LLMs to refine CD-derived features post hoc, offers a promising path toward more robust clinical prediction models. We also find that ICL improves LLM performance in clinical prediction, especially when prompts are tailored to the task and model. Input format also matters: proprietary models prefer RCF, open-source models perform better with JSON, and smaller models benefit from NS inputs. These results highlight the need to align prompts and data formats with model architecture, capacity, and pretraining.

# Appendix

## Contents

# A Reproducibility Statement

## A.1 LLM Implement Details

Our experimental setup utilized 3 NVIDIA RTX A800 GPUs. Our code is built upon the HuggingFace Transformers framework (https://huggingface.co/docs/transformers/en/index). To ensure deterministic outcomes, all Large Language Model (LLM) inferences were performed using Greedy Decoding (i.e., temperature=0, do_sample=False). For full transparency and to facilitate replication, our source code and complete results are publicly available at our project website https://anonymous.4open.science/r/REACT_LLM-5DD1.

Our benchmark evaluates 14 general-purpose LLMs, including Qwen3-8B, Qwen3-14B, Qwen3-235B [37], Llama-3.1-405b [13], DeepSeek-R1 [14], DeepSeek-V3 [24], Gemini-2-Flash, Gemini-2-Pro [28], GPT-4o-mini [26], GPT-o1 [19], GPT-4o [17], Claude-4 [5], GPT-o3-mini [27], Claude-3.5-Haiku [3], Claude-3.7-Sonnet [4]. We obtained the model weights for all publicly available models from the Hugging Face repository (https://huggingface.co/). For the proprietary model commercial models in our benchmark accessible only via API, we specify the official provider and the precise model identifier used for our calls to maintain transparency and facilitate future replication. The specific download links and their corresponding official sources are as follows:

- Qwen3-8B: https://huggingface.co/Qwen/Qwen3-8B
- Qwen3-14B: https://huggingface.co/Qwen/Qwen3-14B
- Qwen3-235B: https://huggingface.co/Qwen/Qwen3-235B-A22B-Instruct-2507
- Llama-3.1-405B: https://huggingface.co/meta-llama/Llama-3.1-405B-Instruct
- DeepSeek-R1: https://huggingface.co/deepseek-ai/DeepSeek-R1-0528
- DeepSeek-V3: https://huggingface.co/deepseek-ai/DeepSeek-V3
- Gemini-2-Pro & Gemini-2-Flash: https://ai.google.dev/models/gemini
- GPT-o1: https://openai.com/zh-Hans-CN/o1/
- GPT-o3-mini: https://platform.openai.com/docs/models/o3-mini
- GPT-4o: https://platform.openai.com/docs/models/chatgpt-4o-latest
- GPT-4o-mini: https://platform.openai.com/docs/models/gpt-4o-mini
- Claude Models (Claude-4, 3.5-Haiku, 3.7-Sonnet): https://docs.anthropic.com/en/docs/about-claude/models/overview

## A.2 Causal Discovery Implement Details

gCastle (https://gcastle.readthedocs.io/en/latest/index.html) is a causal discovery toolkit developed by Huawei Noah's Ark Lab. We utilized its Function-based DirectLiNGAM algorithm, Score-based GES algorithm, and Gradient-based CORL algorithm for our analysis.

The GES algorithm was configured with the parameters `criterion='bic'` and `method='scatter'`. For the CORL algorithm, we used the settings `encoder_name='transformer'`, `decoder_name='lstm'`, `reward_mode='episodic'`, `reward_regression_type='LR'`, `batch_size=len(feature)`, `input_dim=len(feature)`, `embed_dim=64`, and `device_type='gpu'`. The DirectLiNGAM algorithm was applied with its default parameters.

# B Experiment Results

## B.1 Baseline performance - Results of ML Models on MIMIC-III & MIMIC-IV

| Outcome | InHospDeath | | Readmit30 | | MultiICU | | SepsisICU | | AKIICU | | LOS | | EarlyICU | |
|---|---|---|---|---|---|---|---|---|---|---|---|---|---|---|
| **Positive Ratio** | 27.0% | | 14.6% | | 22.0% | | 62.4% | | 47.4% | | 31.3% | | 65.3% | |
| **Metric** | ROC | PRC | ROC | PRC | ROC | PRC | ROC | PRC | ROC | PRC | ROC | PRC | ROC | PRC |
| *Machine Learning Models* | | | | | | | | | | | | | | |
| SVM | 0.9368 | 0.8726 | 0.7120 | 0.2524 | **0.7580** | **0.4992** | 0.8680 | 0.9129 | 0.7778 | 0.7535 | 0.8051 | 0.7035 | 0.6353 | 0.7309 |
| LR | 0.9452 | 0.8836 | **0.7565** | 0.2746 | 0.7533 | 0.4981 | 0.8720 | 0.9137 | 0.7879 | 0.7760 | **0.8269** | **0.7220** | 0.6435 | 0.7616 |
| DT | 0.8029 | 0.7624 | 0.5484 | **0.2747** | 0.5587 | 0.3805 | 0.7416 | 0.8735 | 0.6268 | 0.6971 | 0.6333 | 0.5794 | 0.6190 | 0.8240 |
| RF | 0.9419 | 0.8780 | 0.7084 | 0.2737 | 0.6970 | 0.4068 | **0.8943** | **0.9337** | 0.7967 | 0.7728 | 0.7928 | 0.6707 | 0.7607 | **0.8416** |
| Adaboost | 0.9207 | 0.8373 | 0.7028 | 0.2330 | 0.7146 | 0.4345 | 0.8676 | 0.9143 | 0.7778 | 0.7540 | 0.7673 | 0.6361 | 0.7066 | 0.8137 |
| XGboost | **0.9490** | **0.8843** | 0.7359 | 0.2698 | 0.6985 | 0.4474 | 0.8907 | 0.9212 | **0.8025** | **0.7871** | 0.8127 | 0.6931 | **0.7584** | 0.8412 |
| *Open-source Large Language Models* | | | | | | | | | | | | | | |
| Qwen3-8b | 0.8747 | 0.7458 | 0.4258 | 0.1044 | 0.5477 | 0.2153 | 0.7338 | 0.8356 | 0.7396 | 0.6885 | 0.6072 | 0.3760 | 0.5455 | 0.6738 |
| Qwen3-235b | 0.8807 | 0.7674 | 0.3639 | 0.0944 | 0.5546 | 0.2234 | **0.7608** | **0.8494** | 0.7488 | 0.6962 | 0.5996 | 0.2657 | 0.5603 | 0.7859 |
| DeepSeek-R1 | 0.8860 | 0.7445 | 0.3875 | 0.1021 | 0.5471 | 0.2285 | 0.7341 | 0.8390 | 0.7300 | 0.6924 | 0.5891 | 0.3534 | 0.5189 | 0.7198 |
| *Proprietary Large Language Models* | | | | | | | | | | | | | | |
| Gemini-2-Pro | **0.9167** | **0.8186** | 0.3883 | 0.0984 | 0.5393 | 0.2185 | 0.7568 | 0.8288 | **0.7517** | **0.7296** | 0.6616 | 0.4532 | 0.5016 | 0.6611 |
| Gemini-2-Flash | 0.8975 | 0.7578 | 0.4210 | 0.1044 | 0.5697 | **0.2558** | 0.7267 | 0.8312 | 0.7301 | 0.6792 | 0.6494 | **0.4808** | 0.5128 | 0.7886 |
| GPT-o3-mini | 0.9012 | 0.7581 | **0.5131** | **0.1333** | **0.5800** | 0.2405 | 0.7314 | 0.8334 | 0.7302 | 0.6969 | 0.6098 | 0.3906 | **0.5942** | **0.7888** |

Table 1: Baseline performance of LLMs and traditional ML models across 7 clinical prediction tasks on MIMIC-IV. Bolded scores indicate the best performance within each model category.

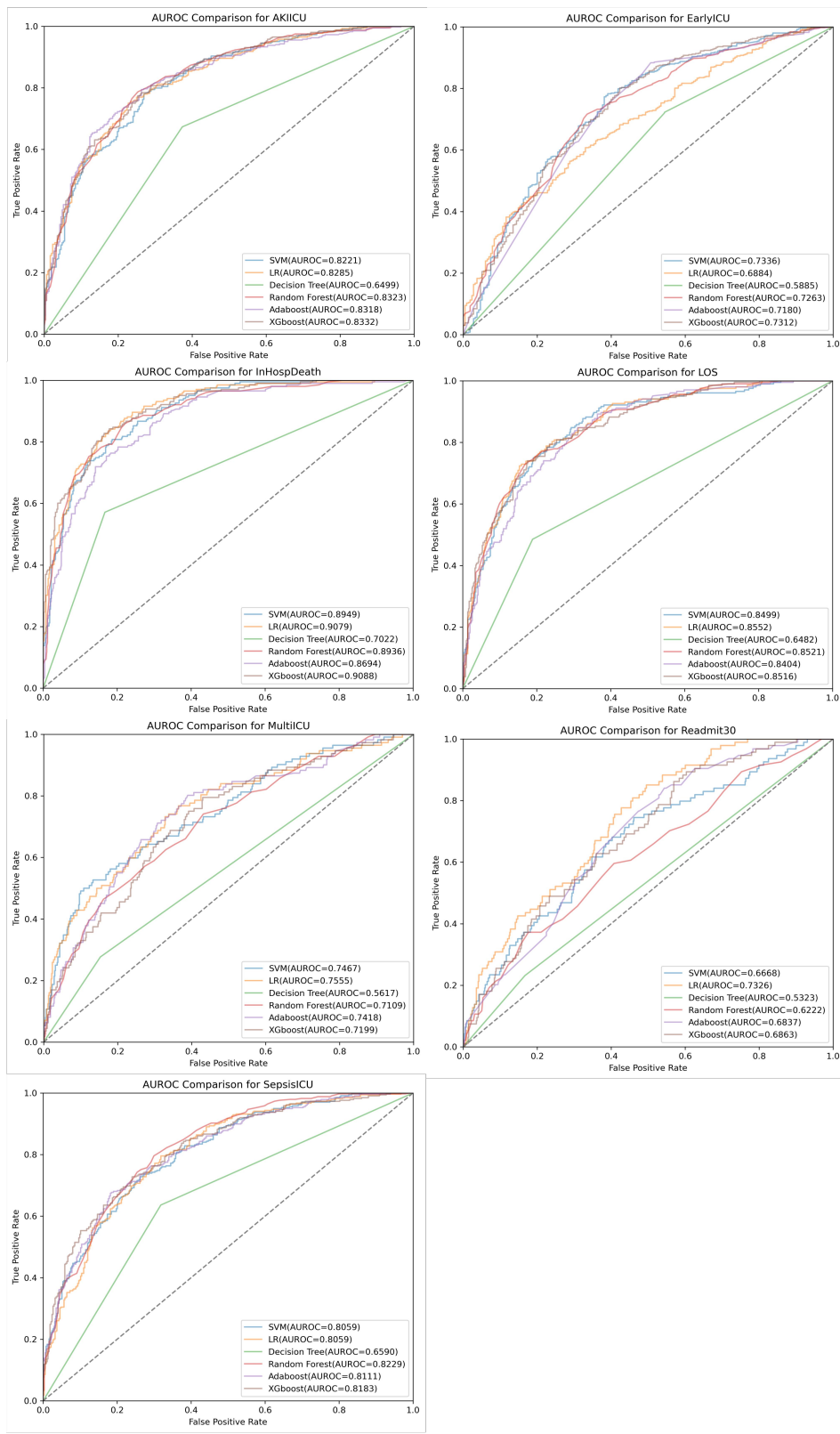

Figure 1: AUROC performance of MLs across 7 clinical prediction tasks on MIMIC-III.

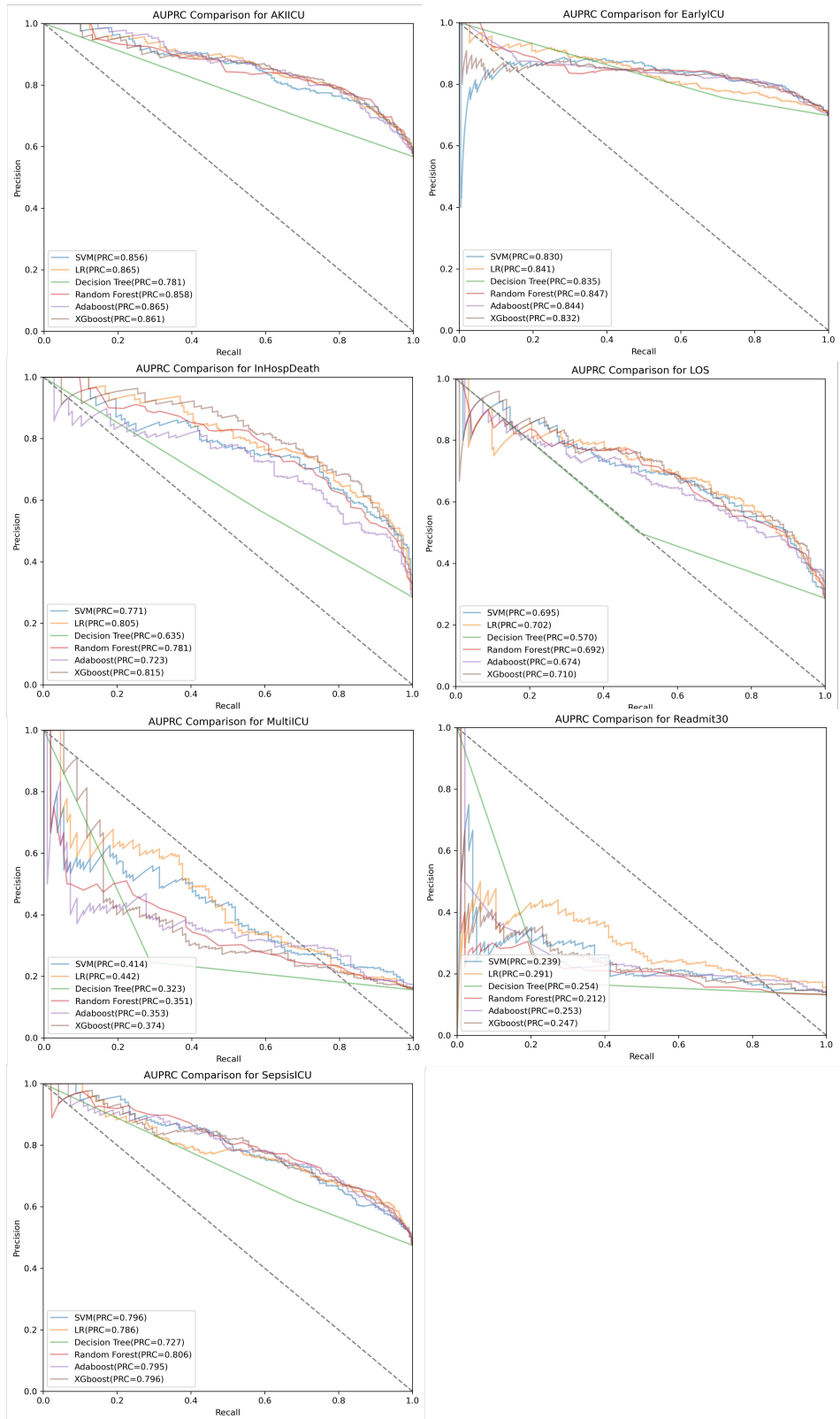

Figure 2: AUPRC performance of MLs across 7 clinical prediction tasks on MIMIC-III.

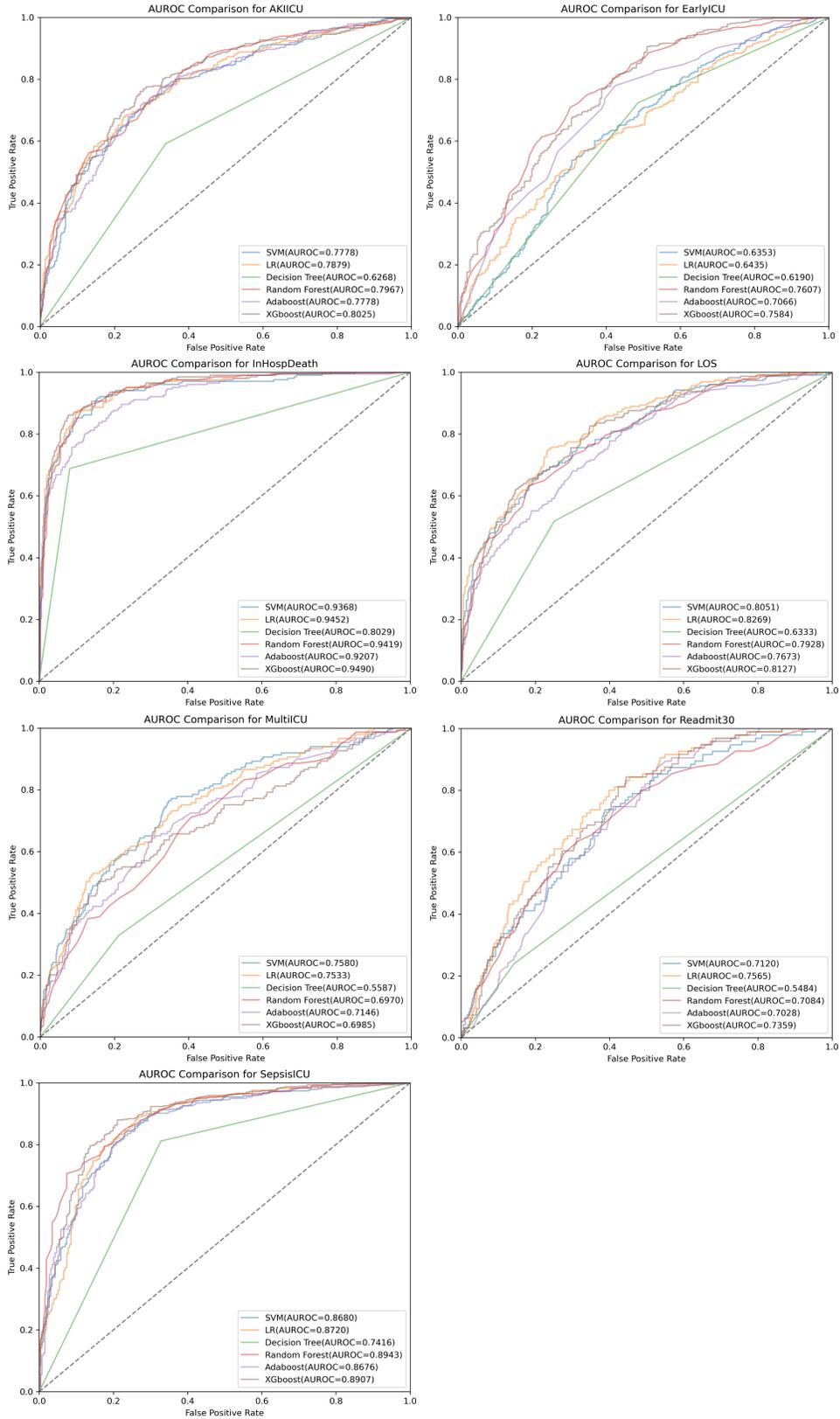

Figure 3: AUROC performance of MLs across 7 clinical prediction tasks on MIMIC-IV.

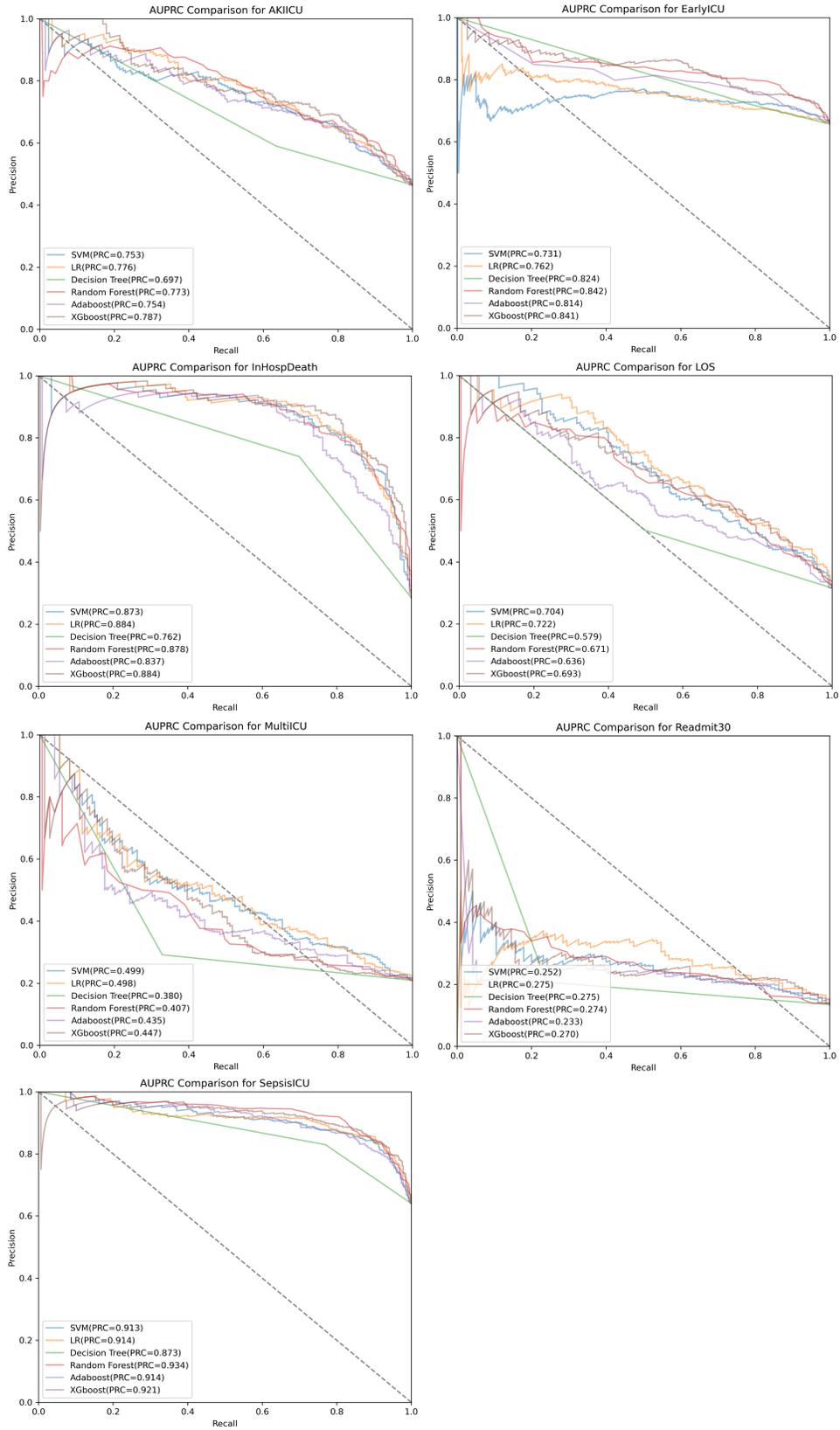

Figure 4: AUPRC performance of MLs across 7 clinical prediction tasks on MIMIC-IV.

## B.2 Baseline performance - Results of Directly Prompting LLMs on MIMIC-III & MIMIC-IV

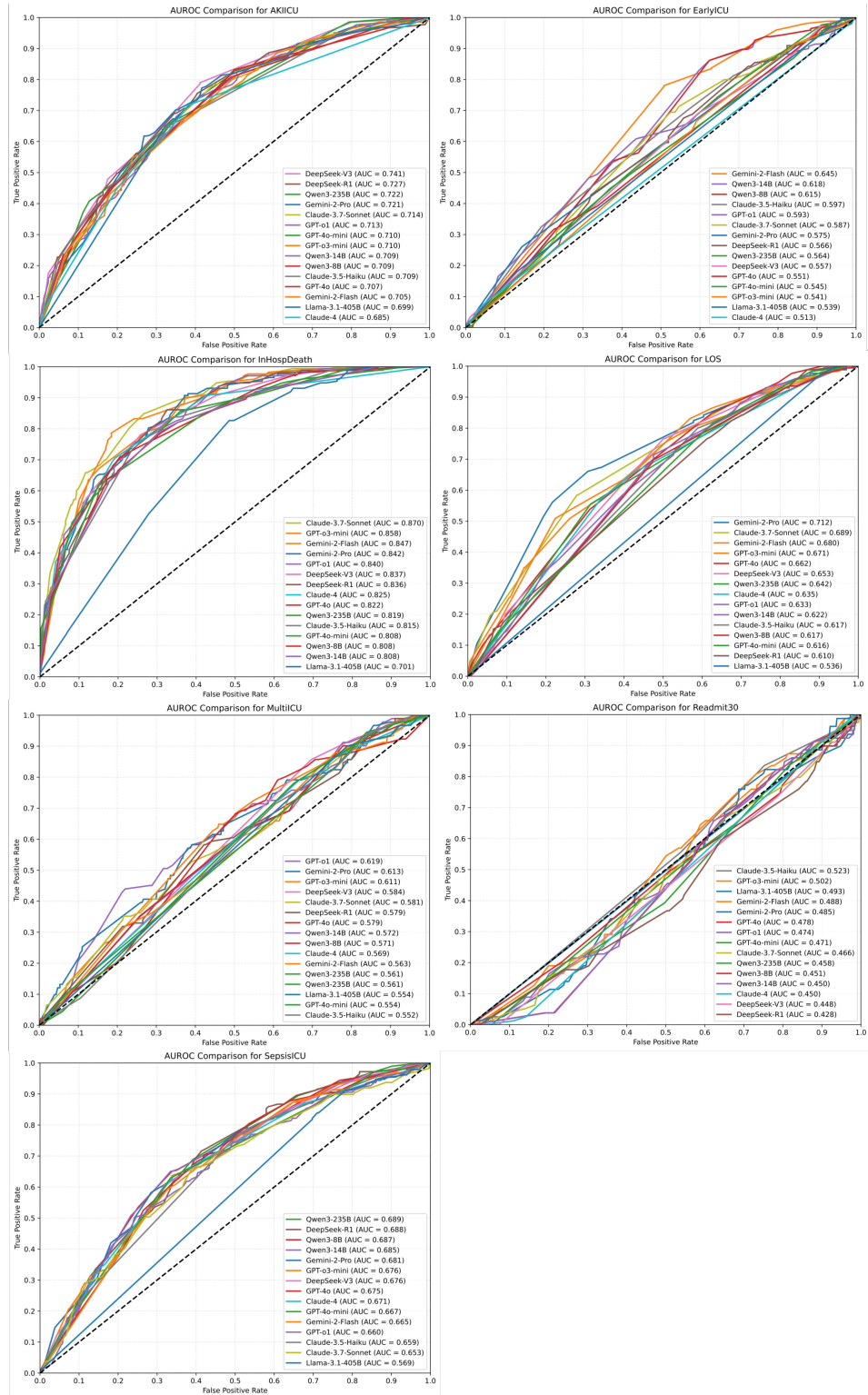

Figure 5: AUROC performance of LLMs across 7 clinical prediction tasks on MIMIC-III.

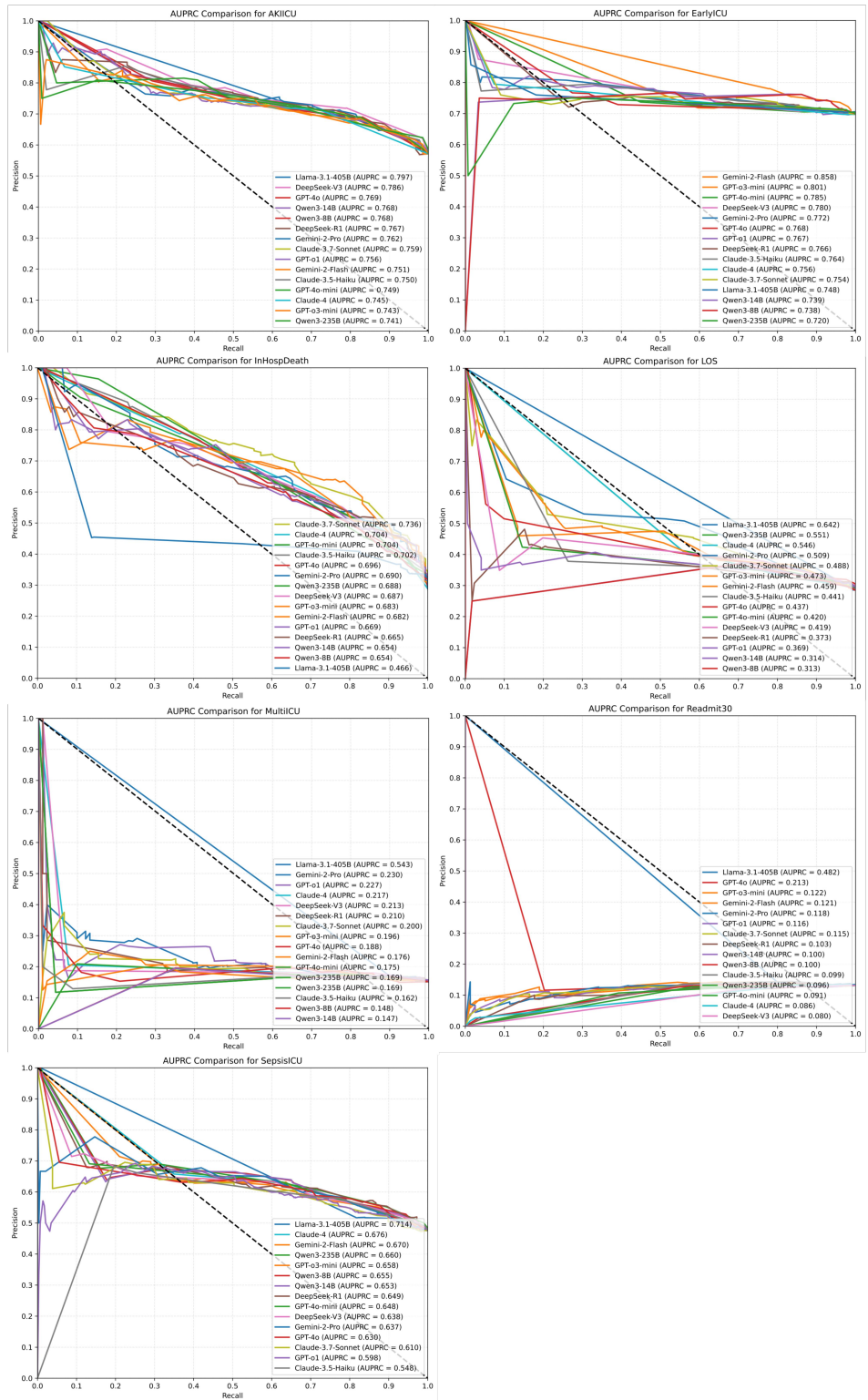

Figure 6: AUPRC performance of LLMs across 7 clinical prediction tasks on MIMIC-III.

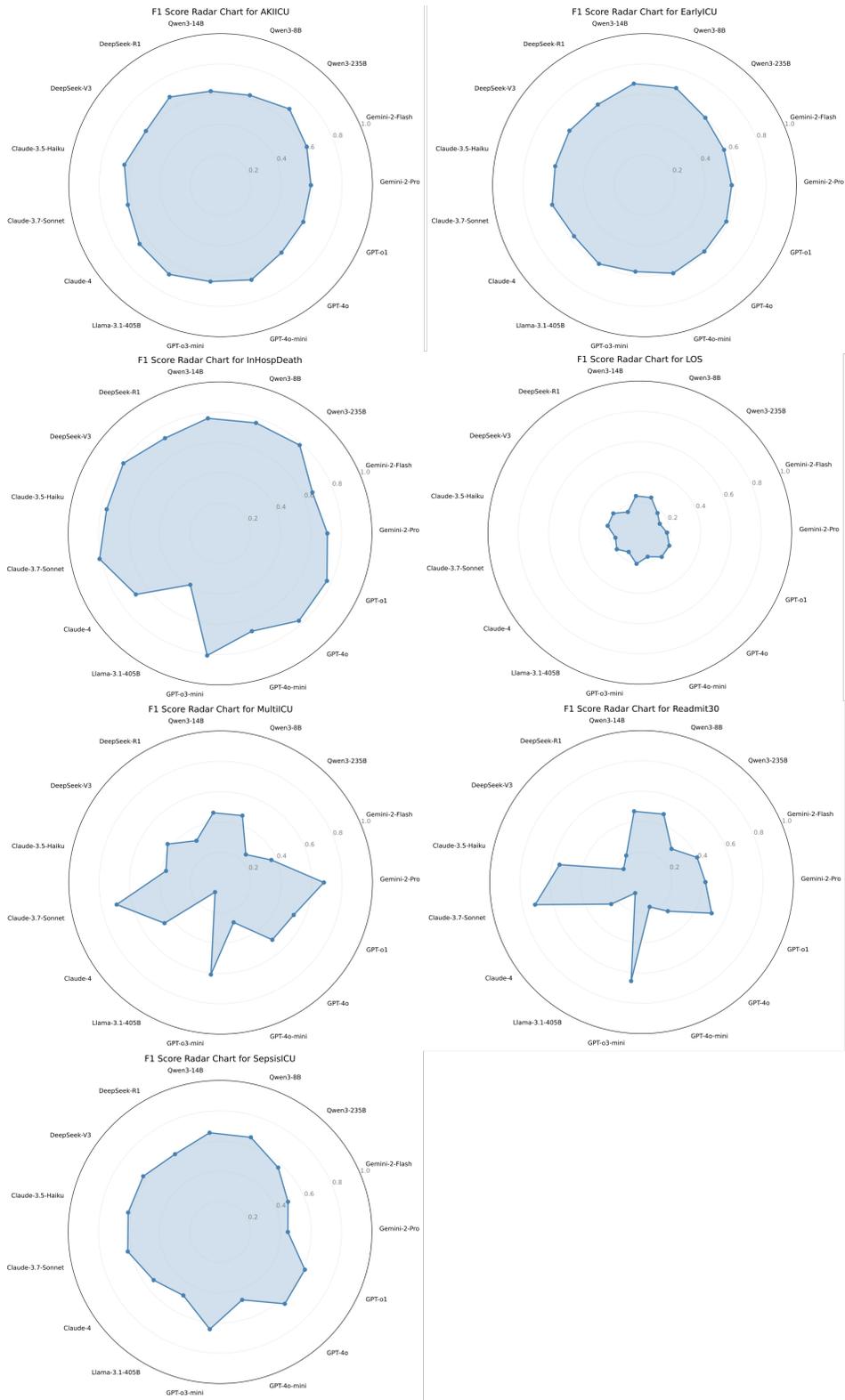

Figure 7: F1 performance of LLMs across 7 clinical prediction tasks on MIMIC-III.

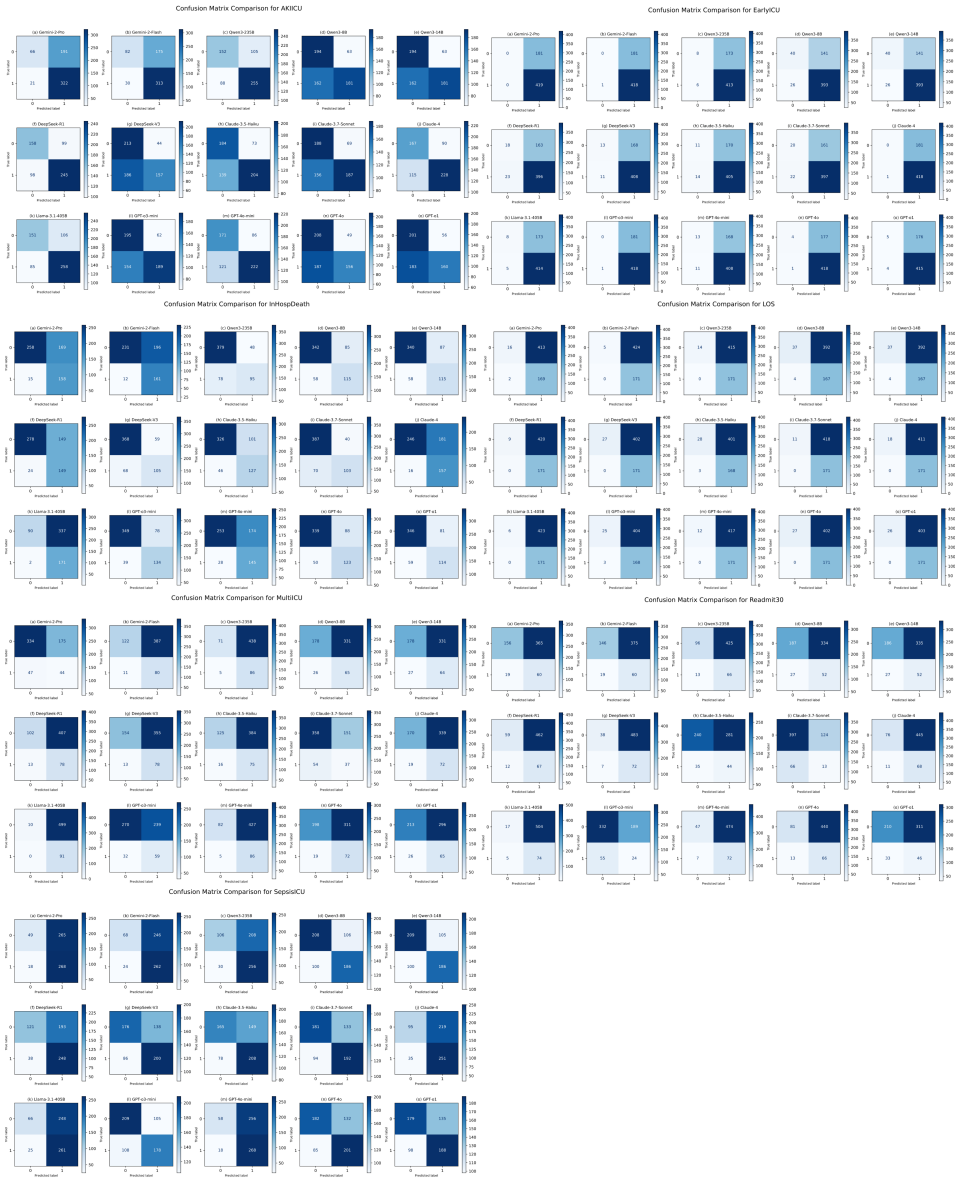

Figure 8: Confusion Matrix performance of LLMs across 7 clinical prediction tasks on MIMIC-III.

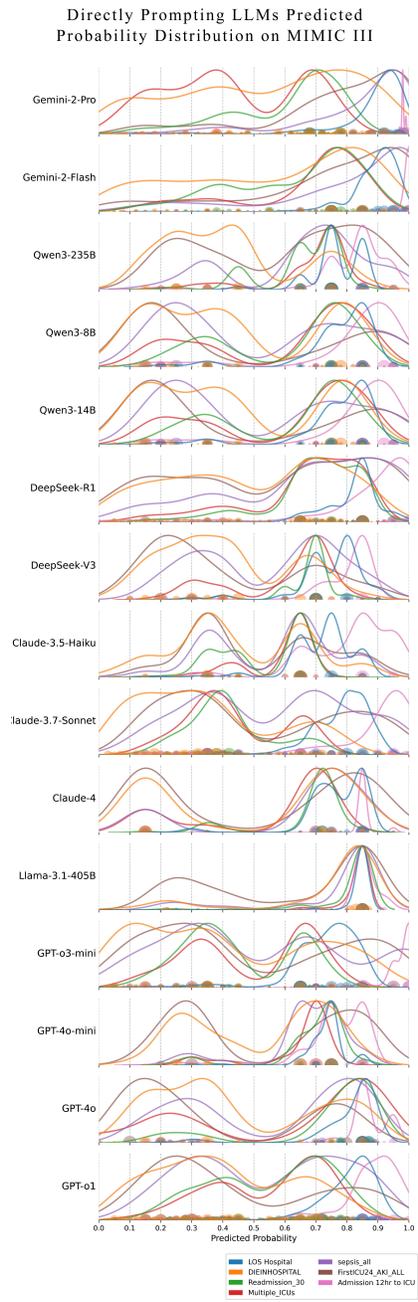

Figure 9: Predictive probability performance of LLMs across 7 clinical prediction tasks on MIMIC-III.

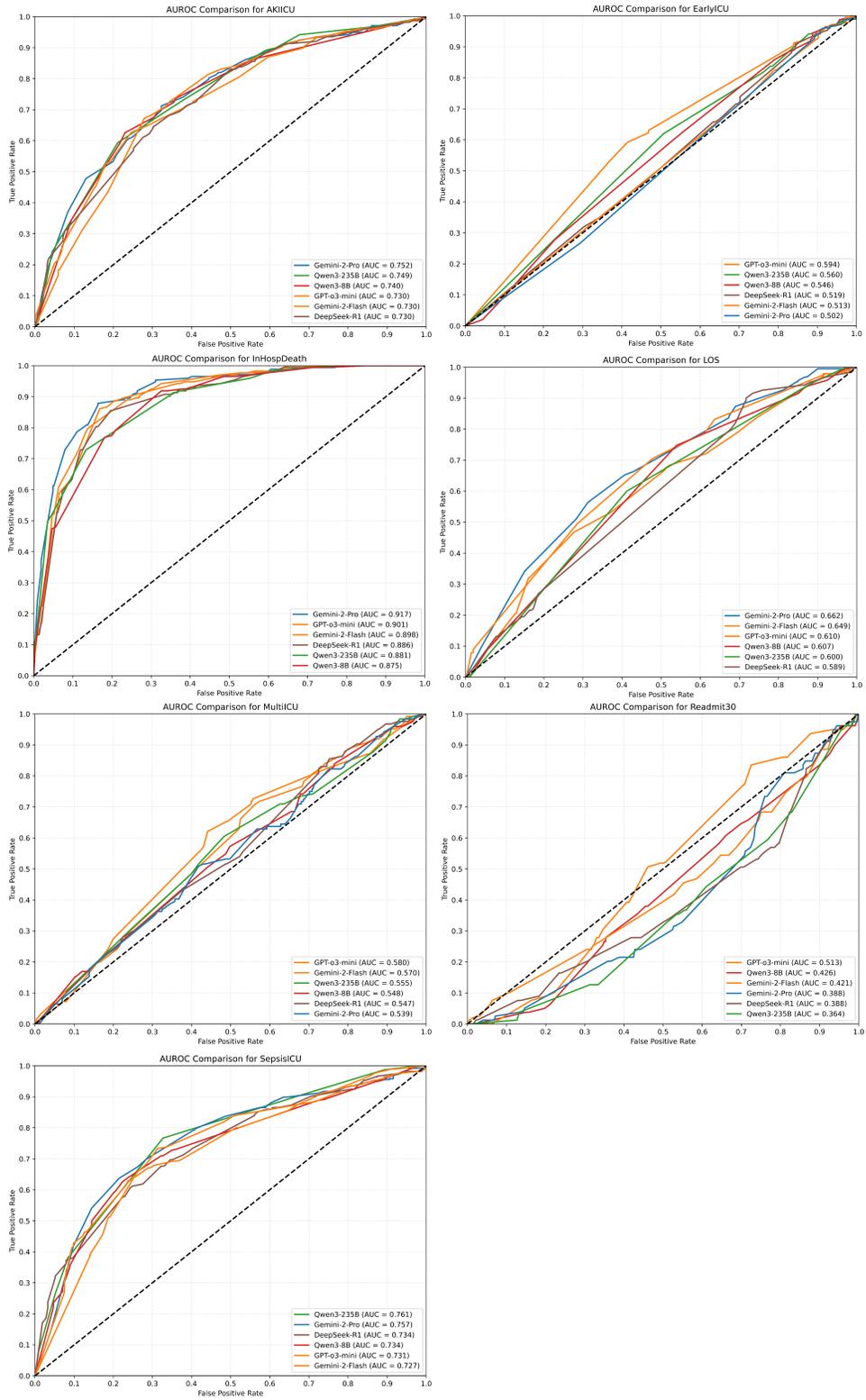

Figure 10: AUROC performance of LLMs across 7 clinical prediction tasks on MIMIC-IV.

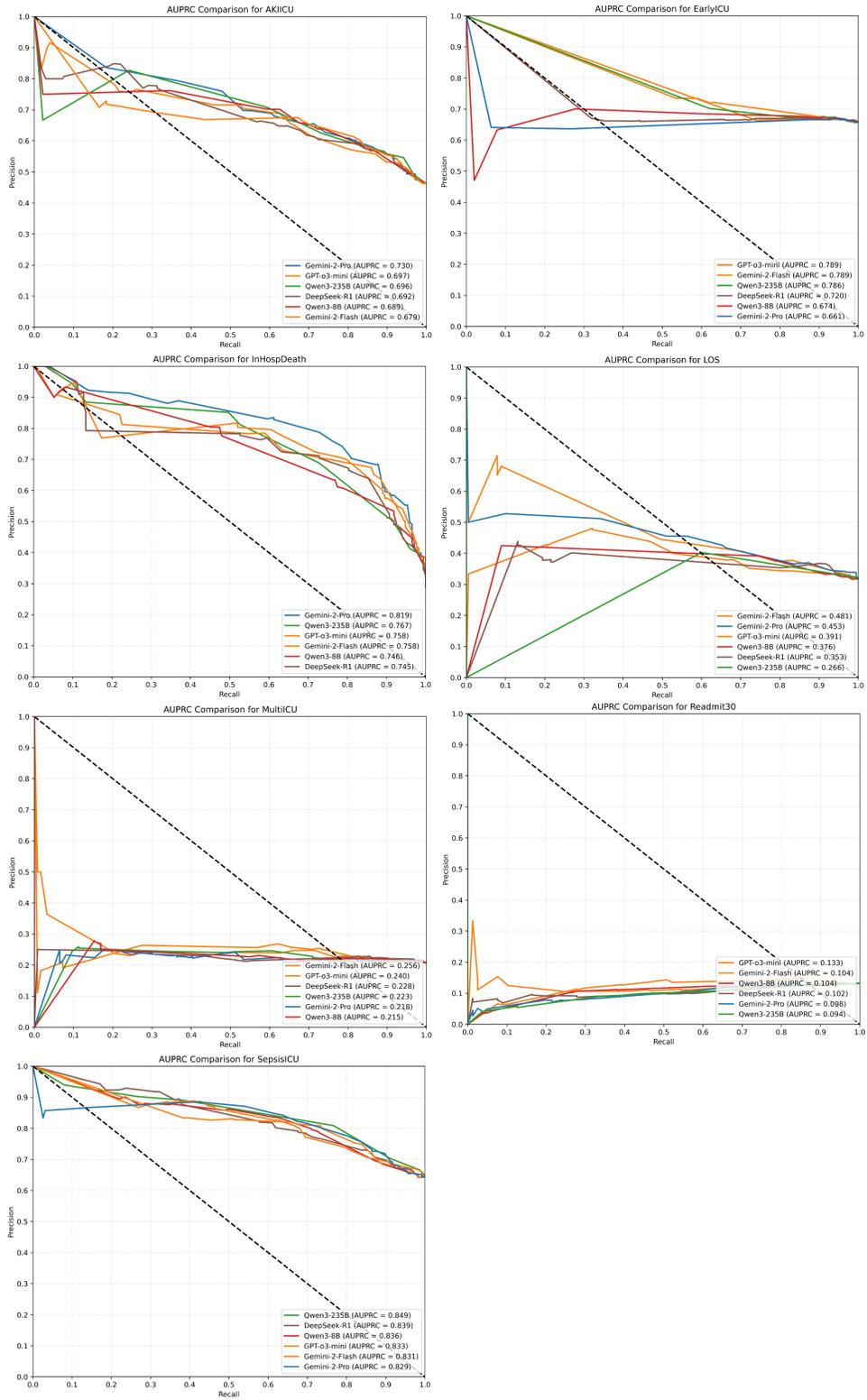

Figure 11: AUPRC performance of LLMs across 7 clinical prediction tasks on MIMIC-IV.

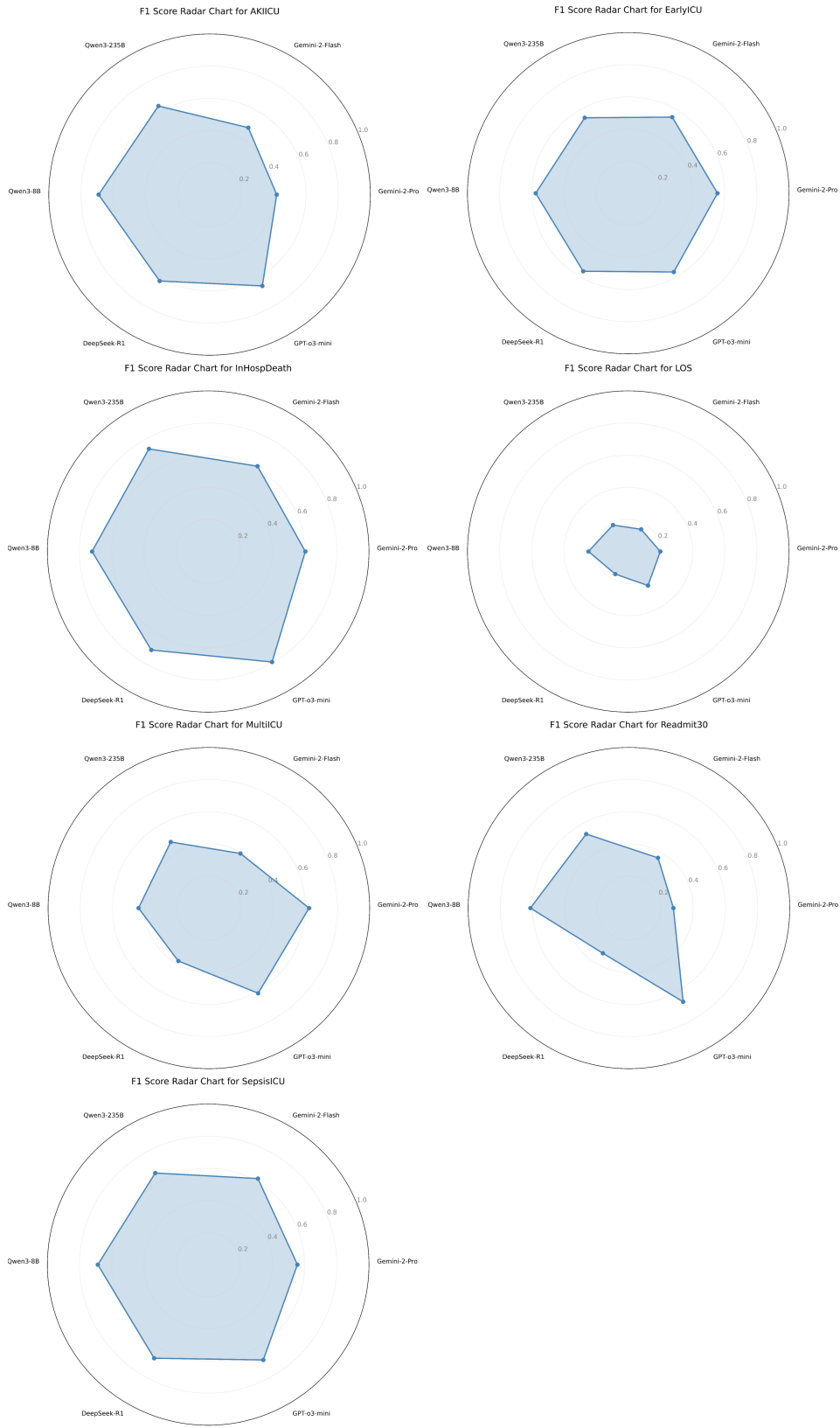

Figure 12: F1 performance of LLMs across 7 clinical prediction tasks on MIMIC-IV.

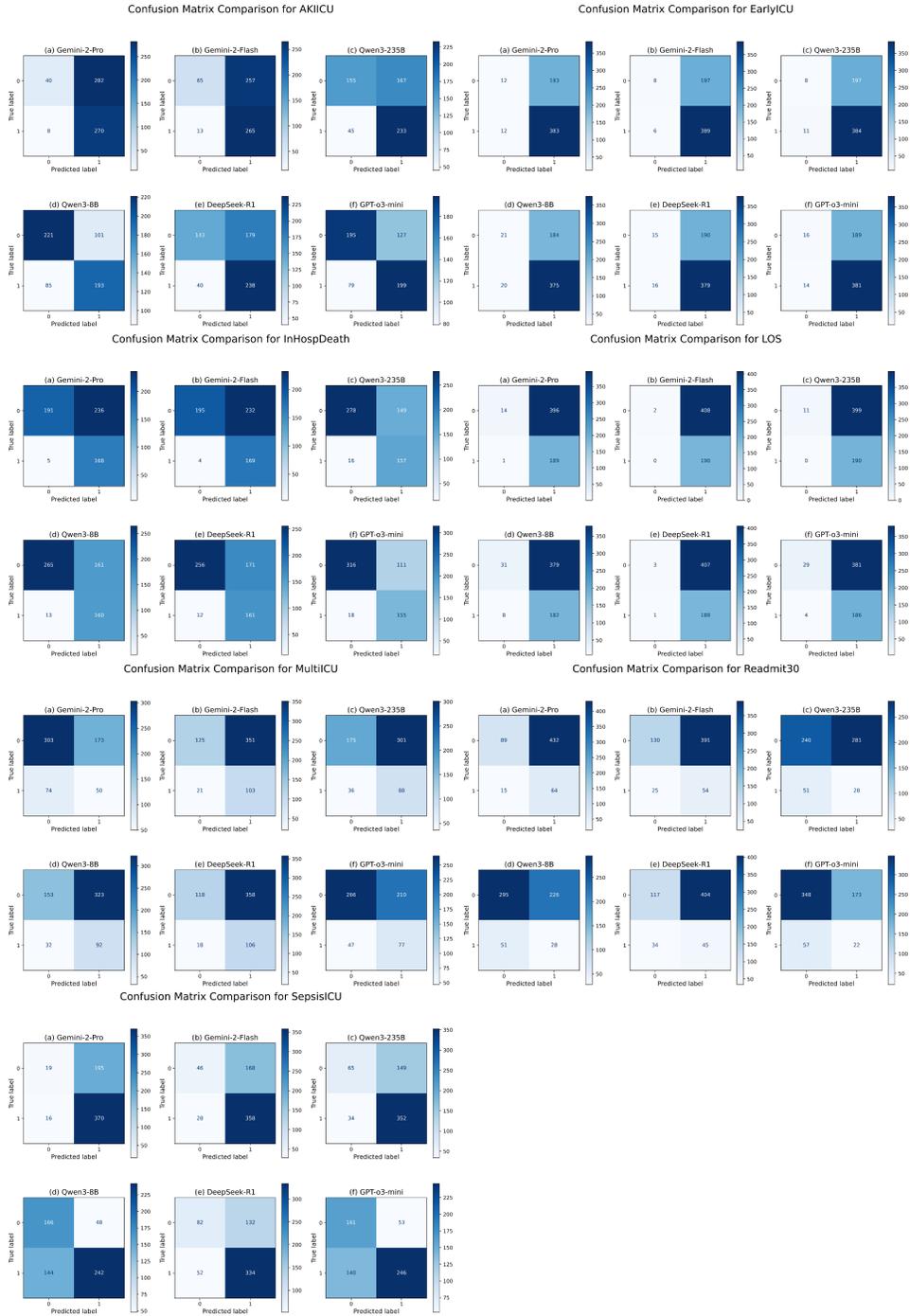

Figure 13: Confusion Matrix performance of LLMs across 7 clinical prediction tasks on MIMIC-IV.

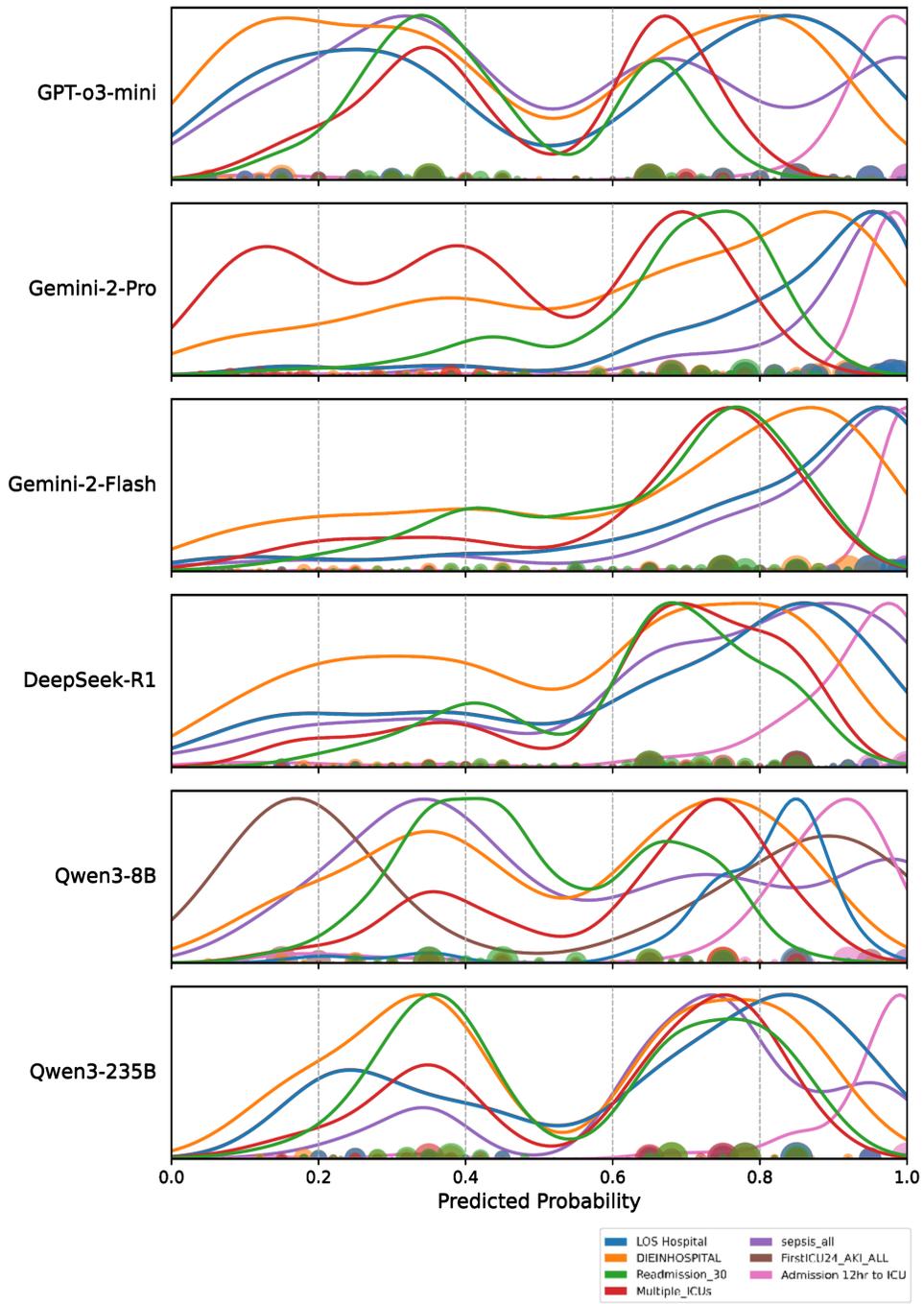

Figure 14: Predictive probability performance of LLMs across 7 clinical prediction tasks on MIMIC-IV.

## B.3 Prompt engineering - Results of Directly LLMs with Prompt engineering on MIMIC-III & MIMIC-IV

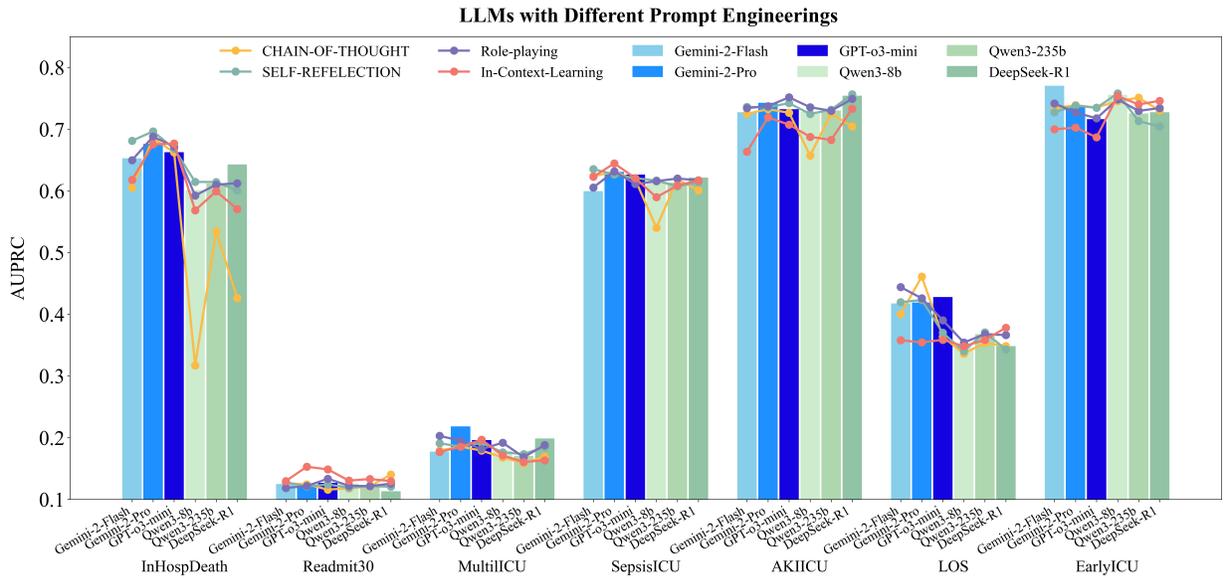

Figure 15: AUPRC of LLMs with different Prompt Engineering on MIMIC-III.

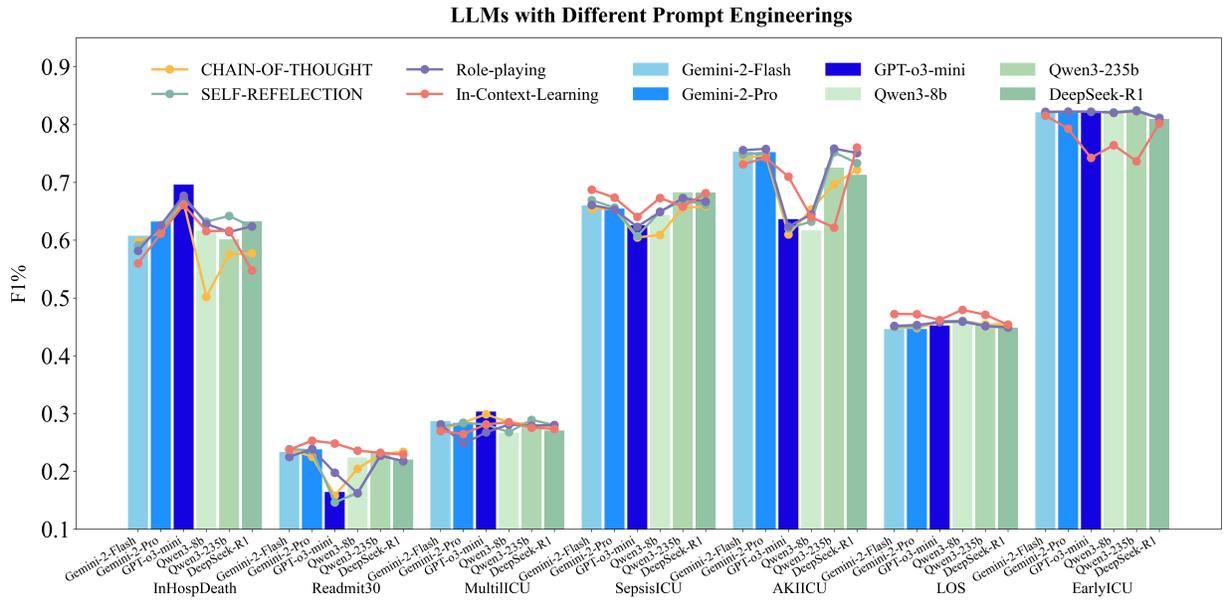

Figure 16: F1 of LLMs with different Prompt Engineering on MIMIC-III.

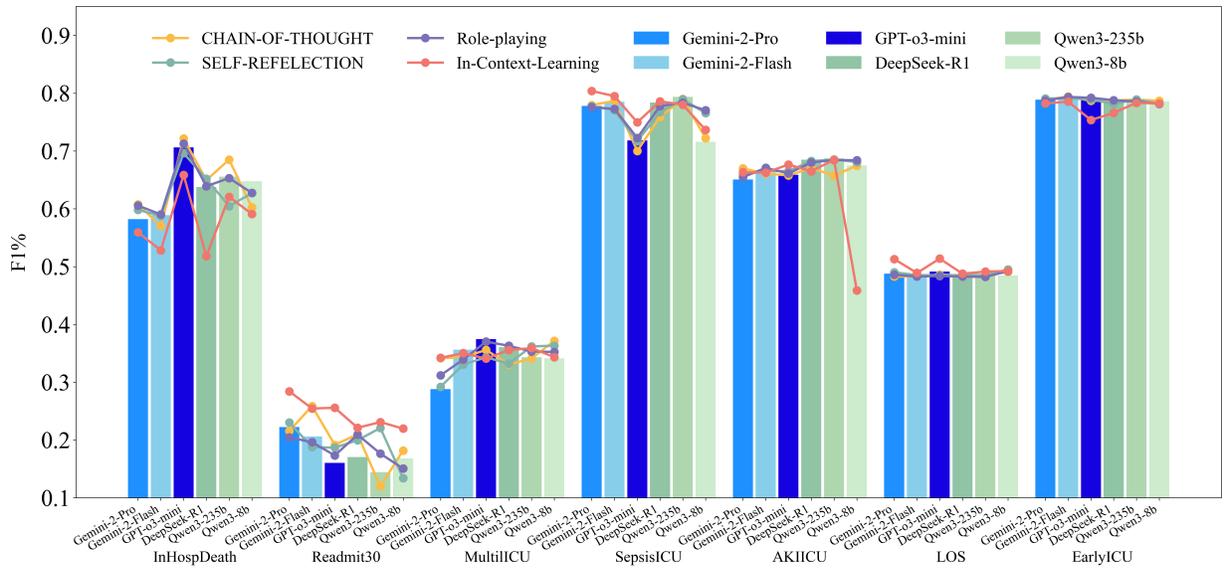

Figure 17: F1 of LLMs with different Prompt Engineering on MIMIC-IV.

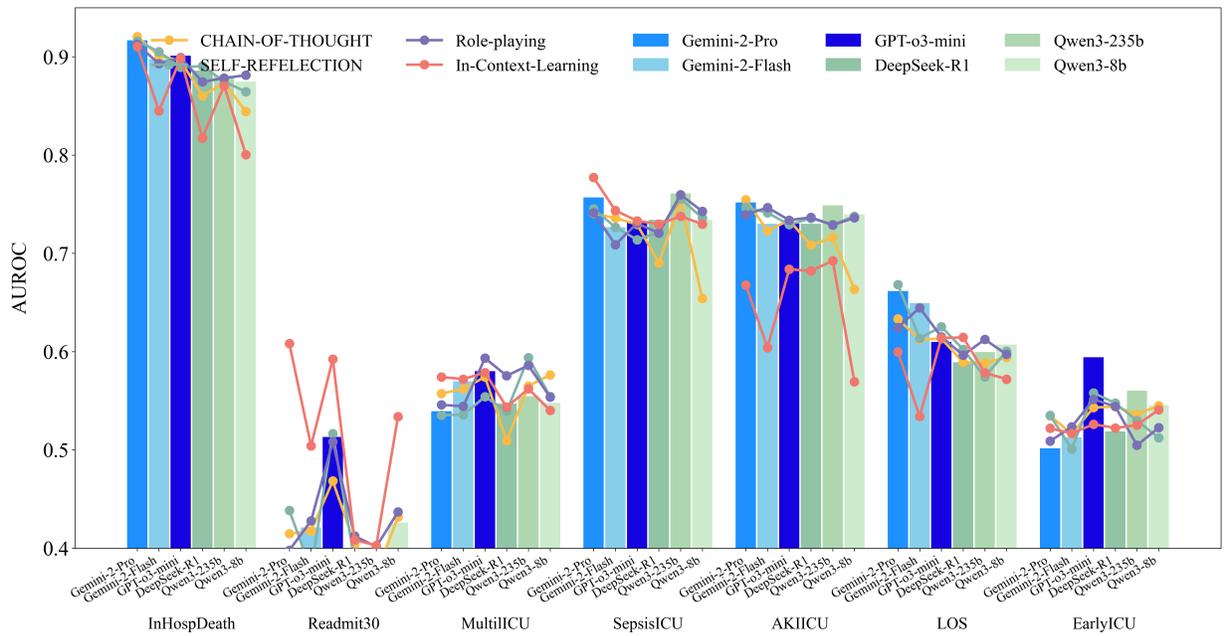

Figure 18: AUROC of LLMs with different Prompt Engineering on MIMIC-IV.

## B.4 Input format sensitivity - Results of Directly Prompting LLMs on MIMIC-III & MIMIC-IV

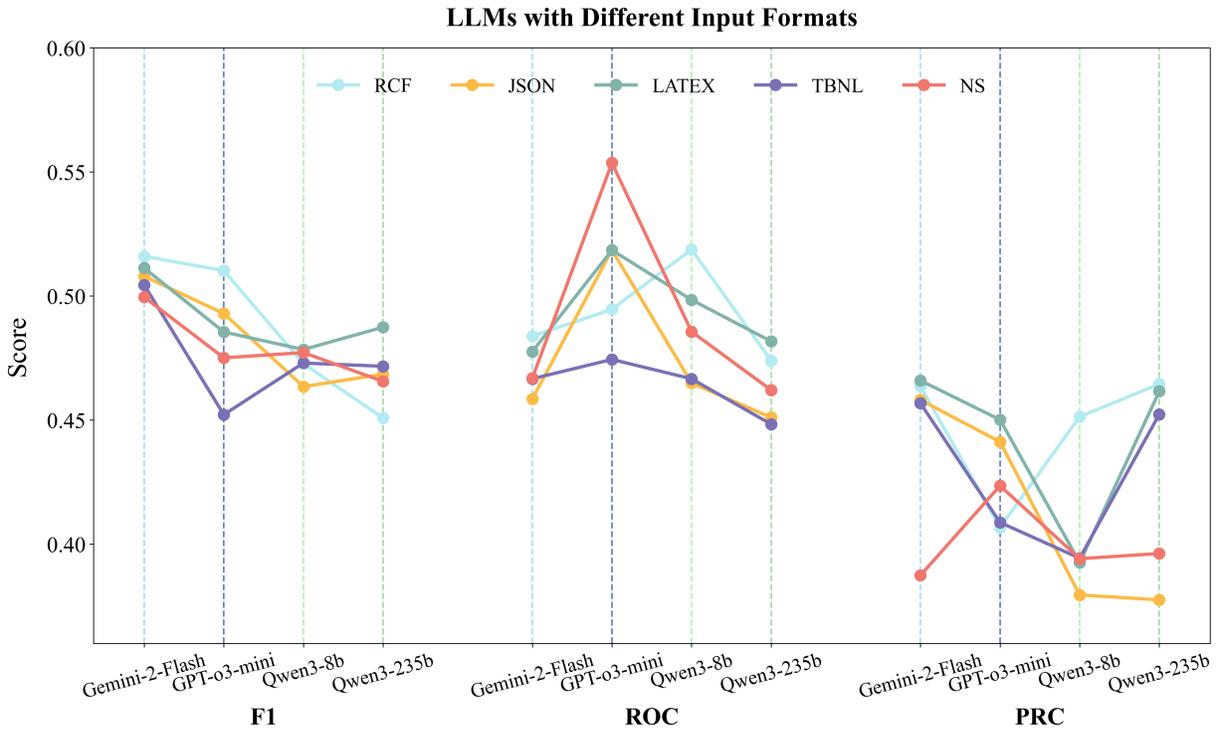

Figure 19: Average Performance of LLMs with Different Input Formats on MIMIC-IV.

## B.5 Causal feature - Causal Discovery Features, LLM-Assisted Causal Features

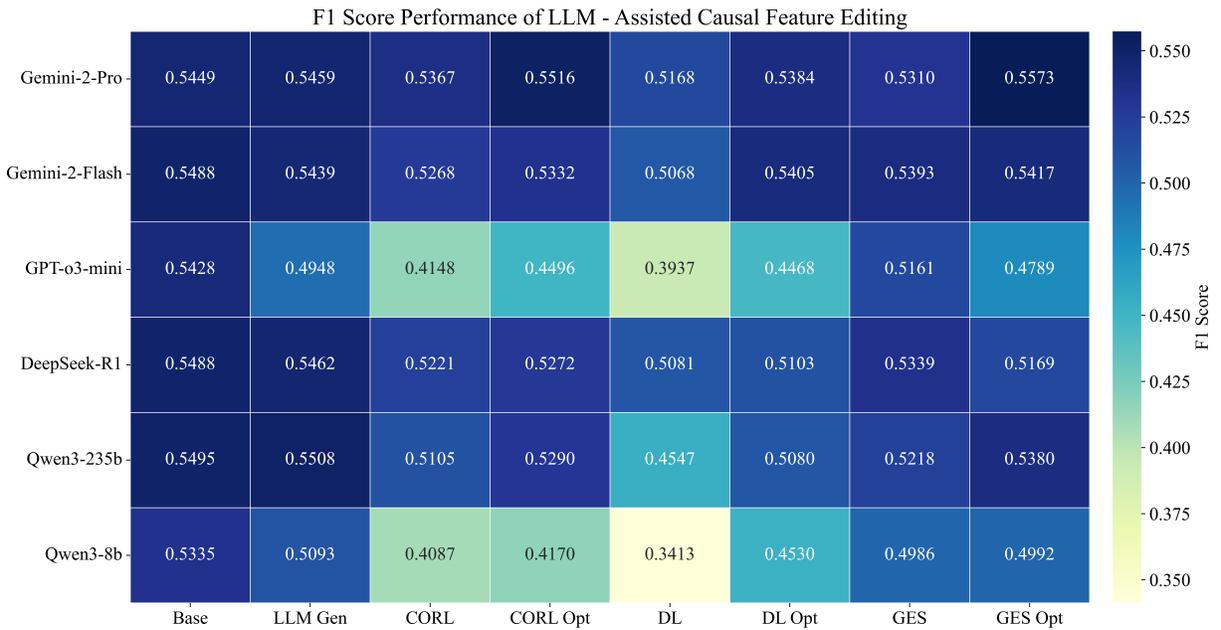

Figure 20: Average F1 of LLMs across 7 outcomes on MIMIC-III and MIMIC-IV.

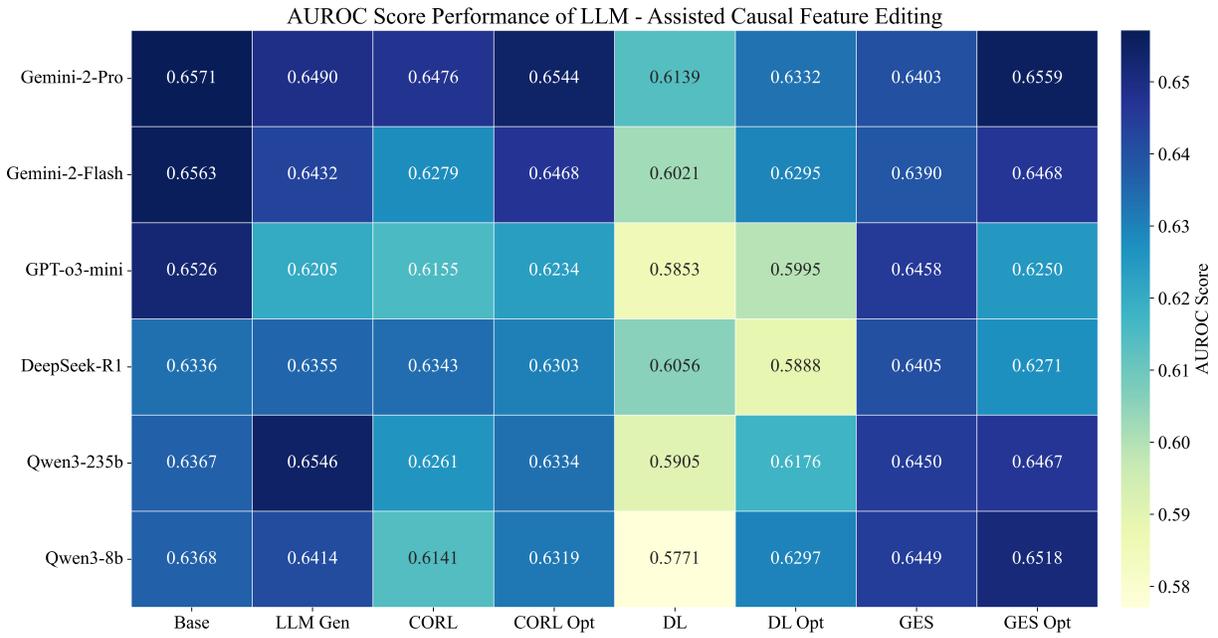

Figure 21: AUROC of LLMs across 7 outcomes on MIMIC-III.

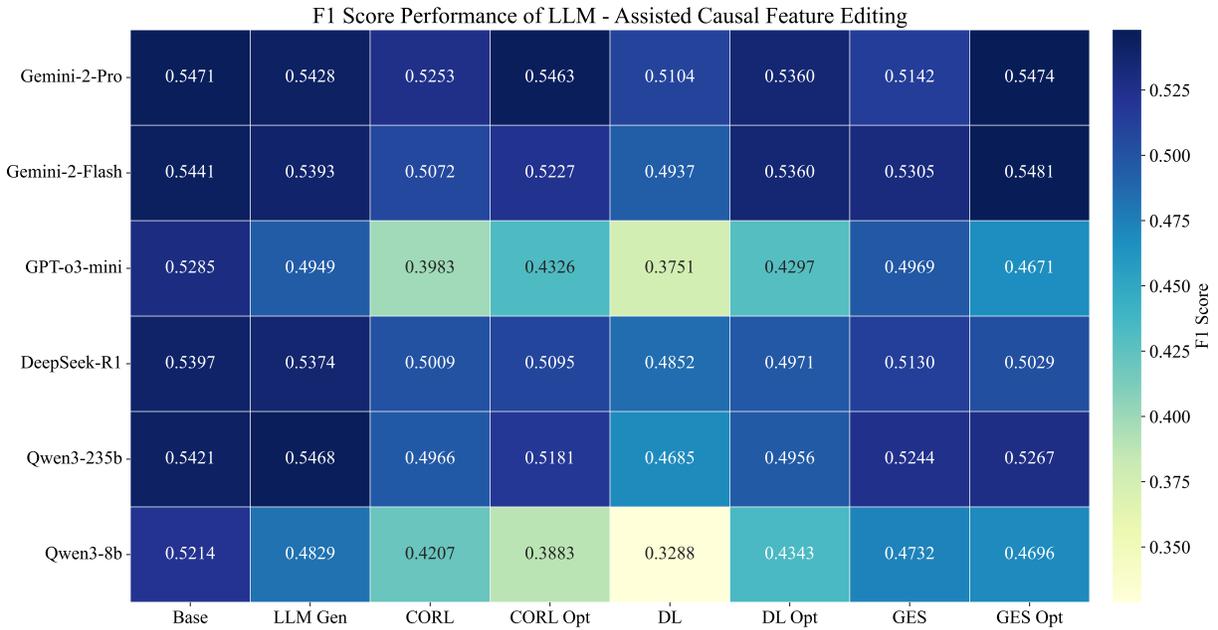

Figure 22: F1 of LLMs across 7 outcomes on MIMIC-III.

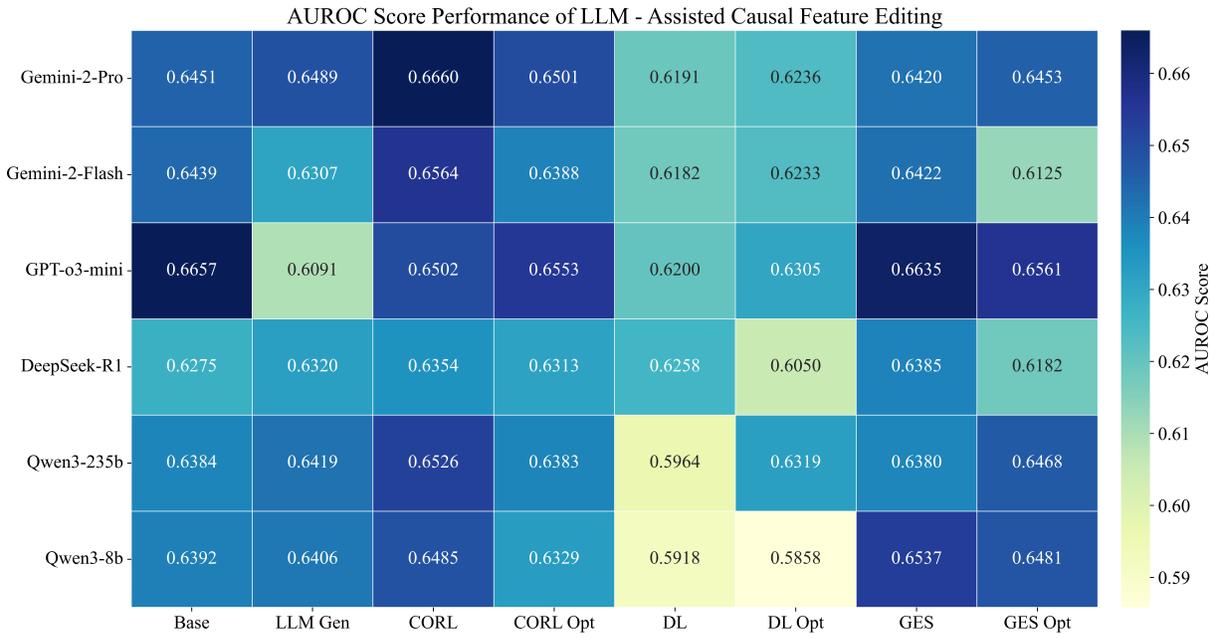

Figure 23: AUROC of LLMs across 7 outcomes on MIMIC-III.

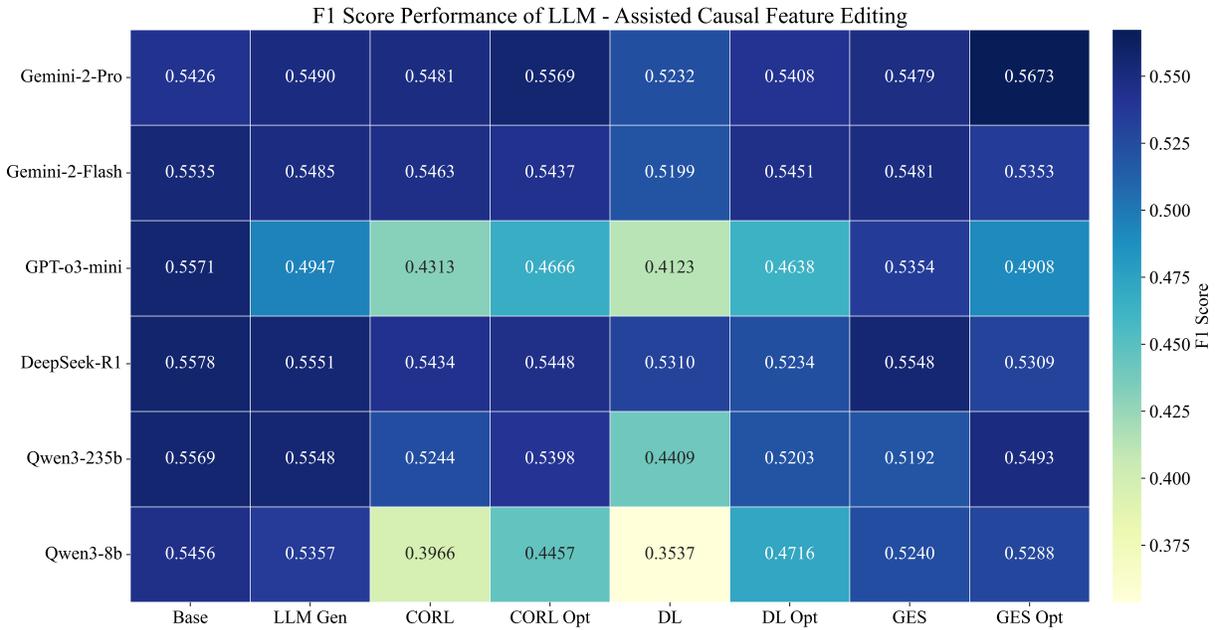

Figure 24: F1 of LLMs across 7 outcomes on MIMIC-III.

# C Examples of Different Input format

## C.1 Narrative Serialization

---

**Example of LLMs with Narrative Serialization for 30-day hospital readmission**

**👤 Patient Basic information**
　Gender: `<Gender>`
　Admission_type: `<Admission_type>`
　First_careunit: `<First_careunit>`
　Age: `<Age>`

**📋 Diagnosis information:** `<Diagnosis feature-list>`

**💉 Procedures information:** `<Procedures feature-list>`

**💊 Medications administered during the first 24 hours after ICU admission:**
　`<Medicine Usage feature-list>`

**🧪 Laboratory test results and vital signs recorded during the first 24 hours after ICU admission:**
　`<TS feature-list>`

> **❓ Your Task:**
> Will the patient die in hospital because of the above situation?
> I know you are not a medical professional, but you are forced to make this prediction.
> Answer with the probability as a number between 0 and 1. Answer with only the number.

---

Figure 25: Prompt template for the Narrative Serialization method. This approach converts structured Electronic Health Record (EHR) data into a human-readable, semi-structured text format for clinical prediction tasks.

## C.2 Row-Column Format

> **Example of LLMs with Row-Column Format for 30-day hospital readmission**
>
> ```
> GENDER, ADMISSION_TYPE, FIRST_CAREUNIT, AGE, Arterial Thrombotic Events, AIDS
>     Diagnosis, Abnormalities of heart beat, ...<other clinical features>
>
> Male, EW EMER., Medical Intensive Care Unit (MICU), 53.0, 1.0, 1, 1.0, ...
>     <other values matching the header structure>
> ```
>
> ❓ **Your Task:**
> Will the patient be readmitted to hospital within 30 days after discharge?
> I know you are not a medical professional, but you are forced to make this prediction.
> Answer with the probability as a number between 0 and 1. Answer with only the number.

Figure 26: Prompt template illustrating the Row-Column serialization format. This method represents patient data in a compact, two-row tabular structure, with the first row defining features and the second containing corresponding values.

## C.3 JSON Encoding

Example of LLMs with JSON Encoding for 30-day hospital readmission

```
{
  "basic_information": {
    "gender": "Male",
    "admission_type": "EW EMER.",
    "first_careunit": "Medical Intensive Care Unit (MICU)",
    "age": 53.0
  },
  "diagnoses": {
    "Arterial Thrombotic Events": {
      "value": 1,
      "original_feature": "Diag_ATE_filtered"
    },
    ...<other blocks of clinical features and values>
  }
}
```

❷ **Your Task:**
Will the patient be readmitted to hospital within 30 days after discharge?
I know you are not a medical professional, but you are forced to make this prediction.
Answer with the probability as a number between 0 and 1. Answer with only the number.

Figure 27: Prompt template for the JSON Encoding method. This format serializes clinical data into a structured JSON object, testing the LLM's ability to parse and reason over machine-readable data structures.

## C.4 LaTeX Table Format

---

**Example of LLMs with LaTeX Table Format for 30-day hospital readmission**

👤 **Patient Basic information**

| GENDER | ADMISSION_TYPE | FIRST_CAREUNIT | AGE |
|--------|----------------|----------------|-----|
| GENDER | ADMISSION_TYPE | FIRST_CAREUNIT | AGE |

📋 **Diagnosis information:**

| Arterial Thrombotic Events | AIDS Diagnosis | Abnormalities of heart beat | ... |
|----------------------------|----------------|-----------------------------|-----|
| 1 | 1 | 1 | ... |

💉 **Procedures information:**

| Fluoroscopy of Multiple Coronary Arteries using Other Contrast | ... |
|----------------------------------------------------------------|-----|
| 1 | ... |

💊 **Medications administered during the first 24 hours after ICU admission:**

| Acetaminophen Usage | Aspirin Usage | Calcium Gluconate Usage | Mean Blood Pressure | ... |
|---------------------|---------------|-------------------------|---------------------|-----|
| <value> | <value> | <value> | <value> | ... |

🩺 **Laboratory test results and vital signs recorded during the first 24 hours after ICU admission:**

| Heart Rate | Systolic Blood Pressure | Diastolic Blood | Mean Blood Pressure | ... |
|------------|-------------------------|-----------------|---------------------|-----|
| <value> | <value> | <value> | <value> | ... |

---

❓ **Your Task:**
Will the patient be readmitted to hospital within 30 days after discharge?
I know you are not a medical professional, but you are forced to make this prediction.
Answer with the probability as a number between 0 and 1. Answer with only the number.

Figure 28: Prompt template for the LaTeX Table serialization method. Patient features are encoded using LaTeX syntax, evaluating the model's proficiency in handling complex markup languages for data interpretation.

## C.5 Template-Based Natural Language

> **Example of LLMs with Template-Based Natural Language for 30-day hospital readmission**
>
> A `<Age>`-year-old `<Gender>` patient was admitted to the `<First_careunit>` as `<Admission_type>`. The patient is diagnosed with `<Diagnosis feature-list containing values>`. The patient has undergone `<Procedure feature-list containing values>`. During the first 24 hours after ICU admission, the patient received `<Medicine Usage feature-list containing values>`. Laboratory tests and vital signs recorded during the first 24 hours showed `<TS feature-list containing values>`.
>
> > ❷ **Your Task:**
> > Will the patient be readmitted to hospital within 30 days after discharge?
> > I know you are not a medical professional, but you are forced to make this prediction.
> > Answer with the probability as a number between 0 and 1. Answer with only the number.

Figure 29: Prompt template for the Template-Based Natural Language approach. This method synthesizes structured EHR data into a cohesive clinical narrative, presenting the information to the LLM in a fluent, paragraph-based format.

# D Examples of LLM-Assisted Causal Feature Editing

## D.1 LLM edited CD feature sets

Examples of GPT-o3-mini edited CD feature sets

 **Input**

You are a clinical expert with extensive experience in intensive care medicine. Your task is to optimize the causal feature sets for various clinical prognostic outcomes generated by the GES causal discovery algorithm, based on established clinical knowledge and clinical experience.

1. All Available Features (`<amount of all features>` features): `<feature-list>`

2. Feature Sets to be Optimized: `<GES-based feature-list>`

3. Your Task and Output Requirements:

Target Outcome Clinical Context: `<label>`:`<description>`

Please consider the specific pathophysiology and clinical risk factors associated with `<label>` when making your optimization decisions.

Complete the following tasks:

1. Add features that you believe are clinically causally related to `<label>` but were missed.

2. Remove features that you believe are not clinically causally related to `<label>`.

3. Return ONLY the changes (additions and removals) in the following JSON format. You must provide clear, concise clinical explanations for each decision.

```
{
  "<label>": {
    "added_features": [
      {
        "feature": "Suggested_Addition_Feature_Description",
        "reason": "Clinical explanation: Why this feature is crucial ..."
      }
    ],
    "removed_features": [
      {
        "feature": "Suggested_Removal_Feature_Description",
        "reason": "Clinical explanation: Why this feature should be ..."
      }
    ]
  }
}
```

Please ensure your response is valid JSON format and includes clinical reasoning for all modifications.

 **Output**

```
{
  "DIEINHOSPITAL": {
    "added_features": [
      {
        "feature": "Age",
        "reason": "Clinical explanation: Age is a well-established risk ..."
      },
      ...
    ],
    "removed_features": [
      {
        "feature": "Metronidazole Usage",
        "reason": "Clinical explanation: Antibiotic usage such as ..."
      },
      ...
    ]
}
```

```
{
  "FirstICU24_AKI_ALL": {
    "added_features": [
      {
        "feature": "Chronic Kidney Disease Group",
        "reason": "Patients with established chronic kidney disease ..."
      },
      ...
    ],
    "removed_features": [
      {
        "feature": "Amiodarone Usage",
        "reason": "Amiodarone is used primarily for arrhythmia management..."
      },
      ...
    ]
}
```

```
{
  "Multiple_ICUs": {
    "added_features": [
      {
        "feature": "Charlson Comorbidity Index",
        "reason": "Clinical explanation: A higher comorbidity burden ..."
      },
      ...
    ],
    "removed_features": [
      {
        "feature": "Vancomycin Usage",
        "reason": "Clinical explanation: Although vancomycin is used in..."
      },
      ...
    ]
}
```

```json
{
  "Readmission_30": {
    "added_features": [
      {
        "feature": "Charlson Comorbidity Index",
        "reason": "Clinical explanation: This index summarizes overall ..."
      },
      ...
    ],
    "removed_features": [
      {
        "feature": "Hereditary factor VIII deficiency",
        "reason": "Clinical explanation: This is a rare condition with ..."
      },
      ...
    ]
  }
}
```

```json
{
  "LOS_Hospital": {
    "added_features": [
      {
        "feature": "White Blood Cell Count",
        "reason": "Clinical explanation: Elevated WBC counts indicate ..."
      },
      ...
    ],
    "removed_features": [
      {
        "feature": "Acetaminophen Usage",
        "reason": "Clinical explanation: As a common symptomatic ..."
      },
      ...
    ]
  }
}
```

```json
{
  "sepsis_all": {
    "added_features": [
      {
        "feature": "White Blood Cell Count",
        "reason": "Clinical explanation: White Blood Cell Count is a key ..."
      },
      ...
    ],
    "removed_features": [
      {
        "feature": "Non-ionized Calcium",
        "reason": "Clinical explanation: While electrolytes can be ..."
      },
      ...
    ]
  }
}
```

```
{
  "ICU_within_12hr_of_admit": {
    "added_features": [
      {
        "feature": "Acute Respiratory Failure Diagnosis",
        "reason": "Patients presenting with acute respiratory failure ..."
      },
      ...
    ],
    "removed_features": [
      {
        "feature": "Multi-Drug Resistant Organisms before ICU",
        "reason": "While important in infection management, a history of ..."
      },
      ...
    ]
}
```

Figure 30: Prompt template for LLM-assisted causal feature set editing. The LLM is instructed to act as a clinical expert to refine a pre-existing feature set (generated by the GES algorithm) by adding or removing features and providing clinical justifications.

### D.2 LLM-Generated causal feature sets

Examples of GPT-o3-mini-Generated causal feature sets for Length-of-hospital-stay

 **Input**

You are a clinical expert with extensive experience in intensive care medicine. Your task is to identify features that have DIRECT or INDIRECT causal relationships with the outcome based on established clinical knowledge and clinical experience.

1. All Available Features (organized by category):
   Diag features ( `<amount of diagnosis features>` features):
   `<Diagnosis feature-list>`
   Proc features ( `<amount of procedure features>` features):
   `<Procedure feature-list>`
   Med features ( `<amount of medicine features>` features):
   `<Medicine Usage feature-list containing values>`
   TS features ( `<amount of TS features>` features): `<TS feature-list containing values>`

2. You need to analyze the above features to identify those that have direct or indirect causal relationships with `<label description>` .

   Consider the following:
   - DIRECT causal relationships: Features that directly cause or strongly predict the outcome
   - INDIRECT causal relationships: Features that are part of the causal pathway, represent underlying pathophysiology, or are established risk factors
   - Include features that reflect disease severity, organ dysfunction, or treatment interventions related to the outcome
   - EXCLUDE features that are purely correlational without established causal basis
   - EXCLUDE features that are consequences rather than causes of the outcome

3. Selection Requirements:
   - Select features that have direct or indirect causal relationships with `<label description>`
   - MINIMUM feature count per category:
   * Diag (Diagnosis): At least 0 features (10% of `<amount of diagnosis features>` )

* Proc (Procedures): At least 0 features (10% of `<amount of procedure features>`)
* Med (Medications): At least 0 features (10% of `<amount of medicine features>`)
* TS (Time Series): At least 0 features (10% of `<amount of TS features>`)
- You may select more than the minimum if clinically relevant

4. Output Requirements:
Return your analysis in the following JSON format with feature descriptions EXACTLY matching those provided in the feature list:

```
{
    "<label>": {
        "selected_features": {
            "Diag": ["feature_description_1", "feature_description_2", ...],
            "Proc": ["feature_description_1", "feature_description_2", ...],
            "Med": ["feature_description_1", "feature_description_2", ...],
            "TS": ["feature_description_1", "feature_description_2", ...],
        },
    },
}
```

IMPORTANT:
- Use EXACT feature descriptions as provided in the feature list
- Organize selections by category (Diag, Proc, Med, TS)
- Meet minimum requirements for each category unless providing valid exemption
- Base selections on established clinical evidence and pathophysiology
- Provide valid JSON format without additional explanations

 Output

```
{
    "DIEINHOSPITAL": {
        "selected_features": {
            "Diag": [
                "AKI Related Group",
                ...
            ],
            "Proc": [
                ""Continuous invasive mechanical ventilation for 96..."
                ...
            ],
            "Med": [
                "Furosemide Usage",
                ...
            ],
            "TS": [
                "Blood Oxygen Saturation",
                ...
            ]
        },
    },
}
```

```json
{
    "Readmission_30": {
        "selected_features": {
            "Diag": [
                "Chronic Kidney Disease Group",
                ...
            ],
            "Proc": [
                "Central venous catheter placement with guidance"
                ...
            ],
            "Med": [
                "Norepinephrine Usage",
                ...
            ],
            "TS": [
                "Heart Rate",
                ...
            ]
        },
    },
}
```

```json
{
    "Multiple_ICUs": {
        "selected_features": {
            "Diag": [
                "Acute Respiratory Failure Diagnosis",
                ...
            ],
            "Proc": [
                "Hemodialysis"
                ...
            ],
            "Med": [
                "Norepinephrine Usage",
                ...
            ],
            "TS": [
                "Heart Rate",
                ...
            ]
        },
    },
}
```

```json
{
    "sepsis_all": {
        "selected_features": {
            "Diag": [
                "Urinary Tract Infection Group",
                ...
            ],
            "Proc": [
                "Central venous catheter placement with guidance"
                ...
            ],
            "Med": [
                "Prednisone Usage",
                ...
            ],
            "TS": [
                "Heart Rate",
                ...
            ]
        },
    },
}
```

```json
{
    "FirstICU24_AKI_ALL": {
        "selected_features": {
            "Diag": [
                "AKI Related Group",
                ...
            ],
            "Proc": [
                "Extracorporeal circulation auxiliary to open heart surgery"
                ...
            ],
            "Med": [
                "Vancomycin Usage",
                ...
            ],
            "TS": [
                "Heart Rate",
                ...
            ]
        },
    },
}
```

```
{
    "LOS_Hospital": {
        "selected_features": {
            "Diag": [
                "Acute Respiratory Failure Diagnosis",
                ...
            ],
            "Proc": [
                "Hemodialysis"
                ...
            ],
            "Med": [
                "Norepinephrine Usage",
                ...
            ],
            "TS": [
                "Heart Rate",
                ...
            ]
        },
    },
}
```

```
{
    "ICU_within_12hr_of_admit": {
        "selected_features": {
            "Diag": [
                "Acute Respiratory Failure Diagnosis",
                ...
            ],
            "Proc": [
                "Central venous catheter placement with guidance"
                ...
            ],
            "Med": [
                "Norepinephrine Usage",
                ...
            ],
            "TS": [
                "Heart Rate",
                ...
            ]
        },
    },
}
```

Figure 31: Prompt template for causal feature generation by an LLM. Acting as a clinical expert, the LLM selects a causally relevant feature set from a comprehensive list of available clinical variables based on provided guidelines.

# E Examples of LLM-Based Prediction

## E.1 Directly Prompting

### E.1.1 Length-of-hospital-stay

---

**Example of LLMs with Directly Prompting for Length-of-hospital-stay**

👤 **Patient Basic information**
   Gender: `<Gender>`
   Admission_type: `<Admission_type>`
   First_careunit: `<First_careunit>`
   Age: `<Age>`

📋 **Diagnosis information:** `<Diagnosis feature-list>`

💉 **Procedures information:** `<Procedures feature-list>`

💊 **Medications administered during the first 24 hours after ICU admission:**
   `<Medicine Usage feature-list>`

🧪 **Laboratory test results and vital signs recorded during the first 24 hours after ICU admission:**
   `<TS feature-list>`

> ❓ **Your Task:**
> Will the patient have a prolonged hospital stay (longer than average)?
> I know you are not a medical professional, but you are forced to make this prediction.
> Answer with the probability as a number between 0 and 1. Answer with only the number.

---

Figure 32: An example of the Direct Prompting technique, here shown for the length-of-stay prediction task. This baseline approach presents serialized patient data followed by a direct query for the outcome.

### E.1.2 ICU admission within 12 hours of hospital admission

> Example of LLMs with Directly Prompting for ICU admission within 12 hours of hospital admission
>
> 👤 **Patient Basic information**
>     Gender: `<Gender>`
>     Admission_type: `<Admission_type>`
>     First_careunit: `<First_careunit>`
>     Age: `<Age>`
>
> 📋 **Diagnosis information:** `<Diagnosis feature-list>`
>
> 💉 **Procedures information:** `<Procedures feature-list>`
>
> 💊 **Medications administered during the first 24 hours after ICU admission:**
>     `<Medicine Usage feature-list>`
>
> 🩺 **Laboratory test results and vital signs recorded during the first 24 hours after ICU admission:**
>     `<TS feature-list>`
>
> > ❓ **Your Task:**
> > Will the patient be admitted to ICU within 12 hours of hospital admission?
> > I know you are not a medical professional, but you are forced to make this prediction.
> > Answer with the probability as a number between 0 and 1. Answer with only the number.

Figure 33: An example of the Direct Prompting technique, in this case applied to predicting ICU admission within 12 hours. This baseline approach presents serialized patient data followed by a direct query for the outcome.

## E.2 Chain-of-Thought Prompting

### E.2.1 Multiple ICU admissions during a single hospitalization

> Example of LLMs with Chain-of-Thought Prompting for Multiple ICU admissions during a single hospitalization
>
> 👤 **Patient Basic information**
>    Gender: `<Gender>`
>    Admission_type: `<Admission_type>`
>    First_careunit: `<First_careunit>`
>    Age: `<Age>`
>
> 📋 **Diagnosis information:** `<Diagnosis feature-list>`
>
> 💉 **Procedures information:** `<Procedures feature-list>`
>
> 💊 **Medications administered during the first 24 hours after ICU admission:**
>    `<Medicine Usage feature-list>`
>
> 🔬 **Laboratory test results and vital signs recorded during the first 24 hours after ICU admission:**
>    `<TS feature-list>`
>
> > ❓ **Your Task:**
> > Will the patient require multiple ICU stays during this hospitalization?
> > I know you are not a medical professional, but you are forced to make this prediction.
> > Please provide your concise reasoning steps for the prediction(no more than 3 steps), and finally answer with the probability as a number between 0 and 1.

Figure 34: An example of the Chain-of-Thought (CoT) prompting strategy, here applied to predicting multiple ICU admissions. This technique instructs the model to provide explicit reasoning steps prior to its final answer.

## E.3 Self-Reflection Prompting

### E.3.1 Sepsis onset during the ICU stay

> **Example of LLMs with Self-Reflection Prompting for Sepsis onset during the ICU stay**
>
> 👤 **Patient Basic information**
>   Gender: `<Gender>`
>   Admission_type: `<Admission_type>`
>   First_careunit: `<First_careunit>`
>   Age: `<Age>`
>
> 📋 **Diagnosis information:** `<Diagnosis feature-list>`
>
> 💉 **Procedures information:** `<Procedures feature-list>`
>
> 💊 **Medications administered during the first 24 hours after ICU admission:**
>   `<Medicine Usage feature-list>`
>
> ✏️ **Laboratory test results and vital signs recorded during the first 24 hours after ICU admission:**
>   `<TS feature-list>`
>
> > ❓ **Your Task:**
> > Will the patient develop sepsis during this hospitalization?
> > I know you are not a medical professional, but you are forced to make this prediction.
> > Answer with the probability as a number between 0 and 1. First answer with a number. Then conduct a concise reflection. Finally output your answer again with a number.

Figure 35: An example of the Self-Reflection prompting technique, in this case for sepsis onset prediction. This method requires the model to generate an initial answer, reflect on its reasoning, and then provide a final, potentially revised, answer.

## E.4 Role-Playing Prompting

### E.4.1 AKI during the ICU stay

---
**Example of LLMs with Role-Playing Prompting for AKI during the ICU stay**

Imagine that you are a doctor. Today, you're seeing a patient with the following profile:

- 👤 **Patient Basic information**
  - Gender: `<Gender>`
  - Admission_type: `<Admission_type>`
  - First_careunit: `<First_careunit>`
  - Age: `<Age>`

- 📋 **Diagnosis information:** `<Diagnosis feature-list>`

- 💉 **Procedures information:** `<Procedures feature-list>`

- 💊 **Medications administered during the first 24 hours after ICU admission:**
  `<Medicine Usage feature-list>`

- 🔬 **Laboratory test results and vital signs recorded during the first 24 hours after ICU admission:**
  `<TS feature-list>`

  > ❓ **Your Task:**
  > Will the patient develop acute kidney injury (AKI) within the first 24 hours of ICU admission?
  > I know you are not a medical professional, but you are forced to make this prediction.
  > Answer with the probability as a number between 0 and 1. Answer with only the number.

---

Figure 36: An example of the Role-Playing prompting technique, here demonstrated for Acute Kidney Injury (AKI) prediction. This approach assigns a specific persona, such as a "doctor," to the LLM to contextualize the prediction task.

## E.5 In-Context Learning

### E.5.1 30-day hospital readmission

---
**Example of In-Context Learning for 30-day hospital readmission**

- 👤 **Patient Basic information**
  - Gender: `Male`
  - Admission_type: `EMERGENCY`
  - First_careunit: `Medical Intensive Care Unit (MICU)`
  - Age: `46.0`

- 📋 **Diagnosis information:** `Arterial Thrombotic Events, AIDS Diagnosis, Abnormalities of heart beat, Hereditary factor VIII deficiency, Mild Liver Disease Diagnosis, Essential (primary) hypertension, Alcoholic liver disease, Other disorders of fluid, electrolyte and acid-base balance`

- 💉 **Procedures information:** `Not available`

- 💊 **Medications administered during the first 24 hours after ICU admission:**
  `Acetaminophen Usage = 1000.0, Aspirin Usage = 650.0, Calcium Gluconate Usage = 8.0, Magnesium Sulfate Usage = 8.0, Norepinephrine Usage = 8.0`

- 🔬 **Laboratory test results and vital signs recorded during the first 24 hours after ICU admission:**

```
Heart Rate=120.0; Systolic Blood Pressure=111.0; Diastolic Blood Pressure=65.0;
Mean Blood Pressure=75.0; Respiratory Rate=25.5; Temperature (Celsius)=37.165;
Oxygen Flow Rate=2.0; ...
```

Answer: 1.0

`<Case 2>`

`<Case 3>`

**👤 Patient Basic information**
   Gender: `<Gender>`
   Admission_type: `<Admission_type>`
   First_careunit: `<First_careunit>`
   Age: `<Age>`

**📋 Diagnosis information:** `<Diagnosis features list>`

**💉 Procedures information:** `<Procedures features list>`

**💊 Medications administered during the first 24 hours after ICU admission:**
   `<Medicine Usage features list>`

**🩺 Laboratory test results and vital signs recorded during the first 24 hours after ICU admission:**
   `<TS features list>`

**❓ Your Task:**
Will the patient be readmitted to hospital within 30 days after discharge?
I know you are not a medical professional, but you are forced to make this prediction.
Answer with the probability as a number between 0 and 1. Answer with only the number.

Figure 37: An example of the In-Context Learning (ICL) strategy, here applied to the 30-day readmission task. This few-shot technique provides the model with several complete examples (case-answer pairs) within the prompt to guide its prediction.

### E.5.2 In-hospital mortality

Example of In-Context Learning for In-hospital mortality

**👤 Patient Basic information**
   Gender: `Male`
   Admission_type: `EMERGENCY`
   First_careunit: `Medical Intensive Care Unit (MICU)`
   Age: `46.0`

**📋 Diagnosis information:** `Arterial Thrombotic Events, AIDS Diagnosis, Abnormalities of heart beat, Hereditary factor VIII deficiency, Mild Liver Disease Diagnosis, Essential (primary) hypertension, Alcoholic liver disease, Other disorders of fluid, electrolyte and acid-base balance`

**💉 Procedures information:** `Not available`

- **Medications administered during the first 24 hours after ICU admission:**
  ```
  Acetaminophen Usage = 1000.0, Aspirin Usage = 650.0, Calcium Gluconate Usage =
   8.0, Magnesium Sulfate Usage = 8.0, Norepinephrine Usage = 8.0
  ```
- **Laboratory test results and vital signs recorded during the first 24 hours after ICU admission:**
  ```
  Heart Rate=120.0; Systolic Blood Pressure=111.0; Diastolic Blood Pressure=65.0;
  Mean Blood Pressure=75.0; Respiratory Rate=25.5; Temperature (Celsius)=37.165;
  Oxygen Flow Rate=2.0; ...
  ```

> **Answer:** 0.0

```
<Case 2>
```
```
<Case 3>
```

- **Patient Basic information**
  - Gender: `<Gender>`
  - Admission_type: `<Admission_type>`
  - First_careunit: `<First_careunit>`
  - Age: `<Age>`

- **Diagnosis information:** `<Diagnosis features list>`

- **Procedures information:** `<Procedures features list>`

- **Medications administered during the first 24 hours after ICU admission:**
  `<Medicine Usage features list>`

- **Laboratory test results and vital signs recorded during the first 24 hours after ICU admission:**
  `<TS features list>`

> **❓ Your Task:**
> Will the patient die in hospital because of the above situation?
> I know you are not a medical professional, but you are forced to make this prediction.
> Answer with the probability as a number between 0 and 1. Answer with only the number.

Figure 38: An example of the In-Context Learning (ICL) strategy, in this case for in-hospital mortality prediction. This few-shot technique provides the model with several complete examples (case-answer pairs) within the prompt to guide its prediction.

## F  Related Work

**Clinical Prediction**: The development of Clinical Decision Support Systems (CDSS) is fueled by the rapid expansion of medical data and the shift toward precision medicine. Prognostic modeling using traditional machine learning (ML) has become mainstream in intensive care units (ICUs), with common tasks including in-hospital mortality [], readmission, acute kidney injury (AKI), and sepsis [12]. These models improve diagnostic efficiency and aid in resource allocation, such as early identification of high-risk patients [18]. Despite rapid progress in deep learning, traditional models—such as logistic regression (LR), XGBoost, and random forests (RF)—remain widely used due to their robustness and interpretability.

**Clinical LLMs**: Recently, large language models (LLMs) have demonstrated potential in clinical prediction and decision-making [36, 15, 34, 1], particularly by directly processing unstructured text and preserving critical clinical information often lost in manual feature extraction. Through fine-tuning or retrieval-augmented generation (RAG), LLMs can integrate hospital-specific data with general medical knowledge, enabling more effective clinical decision support [2, 21]. Most existing LLM benchmarks focus on clinical question-answering tasks [25, 20], multimodal data integration [25, 6], and the development of evaluation frameworks and application scenarios in real-world clinical settings [7, 10].

**Causal learning**: Beyond LLMs, causal learning is also creating new opportunities in clinical prediction by enhancing interpretability at both the model and data levels. For example, in ICU prognosis prediction, models may mistakenly associate spurious variables—such as age and length of hospital stay—due to overfitting or causal inversion, leading to less meaningful predictions [16]. Traditional ML methods often rely on statistical correlations, which can misidentify confounding factors as causal relationships [30, 11]. Causal learning, by contrast, seeks to recover the underlying causal structure from data, typically sampled from conditional distributions. This process, known as causal discovery, involves identifying causal relationships among variables, often represented as a directed acyclic graph (DAG) [40, 38]. Existing causal discovery approaches can be broadly categorized into four groups: (1) constraint-based methods (e.g., PC [33], FCI [9]), (2) score-based methods: these algorithms define a decomposable model score (e.g. BIC) and perform greedy search over equivalence classes to maximize it. The prototypical example is Greedy Equivalence Search (GES) [8] and FGES [29]. (3) functional causal models: by assuming specific structural equation forms or noise distributions, these methods identify causal direction through functional asymmetries. Classic instances are LiNGAM [31] and DirectLiNGAM [32], which exploit non-Gaussianity to fully orient edges in linear SEMs. (4) continuous optimization-based methods: these approaches encode acyclicity as a differentiable constraint and learn the weighted adjacency matrix via gradient descent. Representative examples include NOTEARS and its non-linear extensions, as well as reinforcement-learning-driven ordering models like CORL [35]. Constraint-based methods are not adopted in this study due to two major limitations. First, the faithfulness assumption may not hold in real-world clinical data. Second, these methods cannot distinguish between causal graphs within the same Markov equivalence class, leaving the directionality between variables undetermined.

The integration of LLMs with causal inference has introduced new breakthroughs in clinical prediction, offering reliable causal chains and enhancing interpretability. For example, in cancer gene identification, recent studies have proposed leveraging LLMs to mine vast medical literature for preliminary screening of cancer-related genes. These are then refined using causal inference methods to pinpoint genes with true causal effects. This combined approach enables precise construction of medical causal graphs, improves counterfactual reasoning, and enhances interpretability by translating complex causal structures into clinically meaningful insights [39, 22, 23].